%% file: acl_latex.tex
\pdfoutput=1

\documentclass[11pt]{article}

\usepackage[preprint]{acl}

\usepackage{makecell}
\usepackage{times}
\usepackage{latexsym}
\usepackage{amsmath}
\usepackage{amssymb}
\usepackage{wrapfig}
\usepackage[T1]{fontenc}

\usepackage[utf8]{inputenc}

\usepackage{microtype}

\usepackage{inconsolata}

\usepackage{graphicx}

\usepackage{color,colortbl}
\usepackage{adjustbox}
\definecolor{darkgreen}{rgb}{0,0.5,0}
\definecolor{azureblue}{rgb}{0,0.5,1}
\definecolor{darkgreen}{rgb}{1,0,0}
\definecolor{color1}{HTML}{006EB8}
\definecolor{color2}{HTML}{009B55}
\definecolor{color3}{HTML}{00A99A}
\definecolor{color4}{HTML}{3C8031}
\definecolor{color5}{HTML}{006795}
\definecolor{color6}{HTML}{00AEB3}
\definecolor{mygray}{gray}{0.93}
\definecolor{mygreen}{HTML}{3FBC9D}
\definecolor{arsenic}{rgb}{0.23, 0.27, 0.29}

\usepackage[utf8]{inputenc}
\usepackage[T1]{fontenc}
\usepackage{hyperref}
\hypersetup{
  colorlinks   = true,
  urlcolor     = color1,
  linkcolor    = color1,
  citecolor   = color1
}
\usepackage{url} 
\usepackage{booktabs}
\usepackage{amsfonts}
\usepackage{nicefrac}
\usepackage{microtype}
\usepackage{wrapfig}
\usepackage[nointegrals]{wasysym}
\usepackage{lipsum}  
\usepackage{times}
\usepackage{tikz}
\usepackage{latexsym}
\usepackage{microtype}
\usepackage{graphicx}
\usepackage{placeins}
\usepackage{subcaption}
\usepackage{bbm}
\usepackage{multirow}
\usepackage{enumitem}
\usepackage{tikz}
\usepackage{amsmath}
\usepackage{amssymb}
\usepackage{mathtools}
\usepackage{amsthm}
\usepackage{nccmath}
\usepackage{algorithm}
\usepackage{caption}
\usepackage{listings}
\usepackage[algo2e, ruled]{algorithm2e}
\SetKwComment{Comment}{$\triangleright$\ }{}
\SetKwProg{Init}{Initialize}{}{}

\usepackage{multirow}
\usepackage{pifont}
\usepackage{graphicx} 
\usepackage{xspace}
\newcommand{\pub}[1]{{\color{gray}{\tiny{[{#1}]\!}}}}
\newcommand{\thickhline}{\noalign{\hrule height 1.pt}}

\definecolor{mygray}{gray}{.9}
\definecolor{ggray}{RGB}{127,127,127}
\definecolor{reda}{RGB}{192,0,0}
\definecolor{redb}{RGB}{217,148,143}
\definecolor{myyellow}{RGB}{255,220,120}
\definecolor{mygreen}{RGB}{80,100,40}
\definecolor{myblue}{RGB}{30,90,100}

\usepackage[percent]{overpic}
\usepackage{fontawesome}
\usepackage{seqsplit}

\definecolor{mygreen}{RGB}{80,100,40}
\definecolor{myblue}{RGB}{30,90,100}

\definecolor{color2}{HTML}{009B55}
\newcommand{\brc}{\textcolor{color2}}

\title{To See a World in a Spark of Neuron: Disentangling \\Multi-task Interference for Training-free Model Merging}

\author{
 \textbf{Zitao Fang\textsuperscript{1}} \quad
 \textbf{Guodong Du\textsuperscript{2,*}} \quad
 \textbf{Shuyang Yu\textsuperscript{3,\dag}} \quad
 \textbf{Yifei Guo\textsuperscript{4,\dag}}
\\
 \textbf{Yiwei Zhang\textsuperscript{1}} \quad
 \textbf{Yiyao Cao\textsuperscript{1}} \quad
 \textbf{Jing Li\textsuperscript{5}} \quad
 \textbf{Ho-Kin Tang\textsuperscript{5}} \quad
 \textbf{Sim Kuan Goh\textsuperscript{1,*}}
\\
 \textsuperscript{1}Xiamen University Malaysia \quad
 \textsuperscript{2}The Hong Kong Polytechnic University
\\
 \textsuperscript{3}Columbia University \quad
 \textsuperscript{4}Duke University \quad
\textsuperscript{5}Harbin Institute of Technology (Shenzhen)
\\
 \small{
   \href{mailto:ait2209071@xmu.edu.my}{ait2209071@xmu.edu.my}, 
   \href{mailto:duguodong7@gmail.com}{duguodong7@gmail.com}, 
   \href{mailto:simkuangoh@gmail.com}{simkuangoh@gmail.com}
 }
}

\begin{document}
\maketitle

\begingroup
\renewcommand\thefootnote{\fnsymbol{footnote}}
\footnotetext[1]{Corresponding authors.}
\footnotetext[2]{Work conducted while at Xiamen University Malaysia.}
\endgroup

\input{sections/0_Abstract}
\input{sections/1_Introduction}
\input{sections/2_RelatedWork}
\input{sections/3_Methodology}
\input{sections/4_ExperimentalSetup}
\input{sections/5_Results}
\input{sections/6_AdditionalResults}
\input{sections/7_Conclusion}
\clearpage
\input{sections/8_Limitations}

\newpage
\bibliography{custom}

\input{sections/9_Appendix}

\end{document}

%% file: sections/0_Abstract.tex
\begin{abstract}
Fine-tuning pre-trained models on targeted datasets enhances task-specific performance but often comes at the expense of generalization. Model merging techniques, which integrate multiple fine-tuned models into a single multi-task model through task arithmetic, offer a promising solution. However, task interference remains a fundamental challenge, leading to performance degradation and suboptimal merged models. Existing approaches largely overlooked the fundamental roles of neurons, their connectivity, and activation, resulting in a merging process and a merged model that does not consider how neurons relay and process information. In this work, we present the first study that relies on neuronal mechanisms for model merging. Specifically, we decomposed task-specific representations into two complementary neuronal subspaces that regulate input sensitivity and task adaptability. Leveraging this decomposition, we introduced \mbox{\textbf{NeuroMerging}}, a novel merging framework developed to mitigate task interference within neuronal subspaces, enabling training-free model fusion across diverse tasks. Through extensive experiments, we demonstrated that \mbox{NeuroMerging} achieved superior performance compared to existing methods on multi-task benchmarks across both natural language and vision domains. Our findings highlighted the importance of aligning neuronal mechanisms in model merging, offering new insights into mitigating task interference and improving knowledge fusion. Our project is available at \url{https://ZzzitaoFang.github.io/projects/NeuroMerging/}.
\end{abstract}

%% file: sections/1_Introduction.tex
\section{Introduction}\label{Introduction}

{\hspace*{1.6em}"To see a world in a grain of sand..."}

\vspace{0.3em}
{\small \rightline{\textit{--William Blake}}}
\vspace{0.5em}

\begin{figure}[t]
  \includegraphics[width=\columnwidth, trim=20 91 386 94, clip]{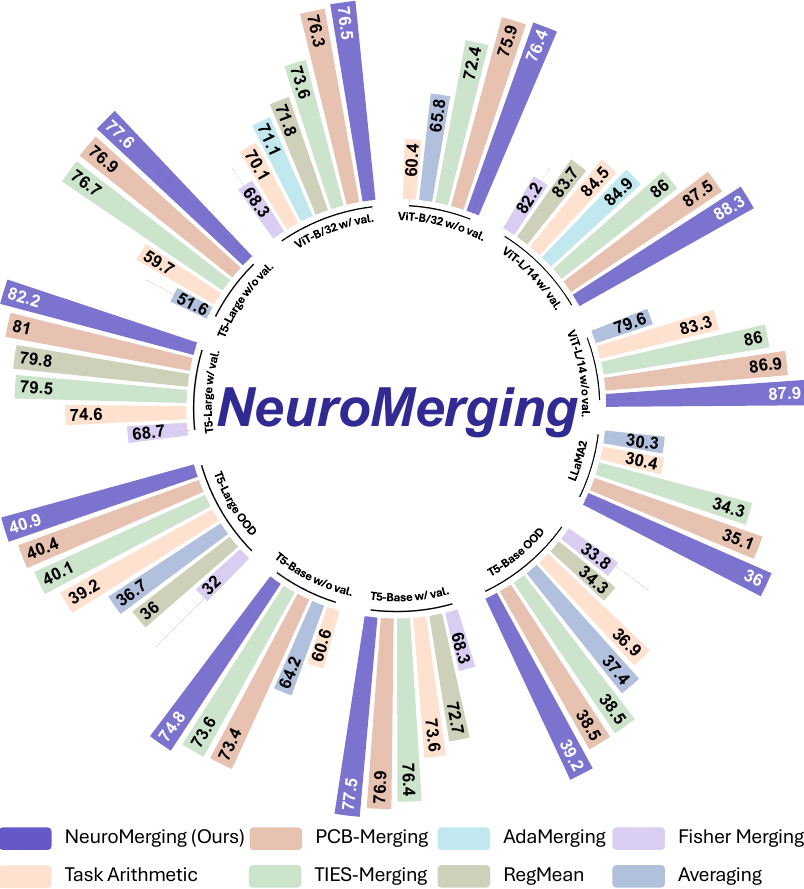}
  \caption{Comparative analysis of our method with baselines across models and domains.}
  \label{fig:intro-sun}
\end{figure}

\input{tables/TVvsNTV}

\input{tables/intro_compare}

\noindent To see a world in the spark of a neuron, the essence of deep learning lies in the complex dynamics of interconnected neurons, which has been empowering recent pre-trained models (PTMs), such as foundation models and large language models (LLMs)~\cite{vaswani2017attention,achiam2023gpt,touvron2023llama}, to learn rich representations from large-scale datasets. These models demonstrate general capabilities while enabling effective fine-tuning for task-specific adaptation~\cite{touvron2023llama}. PTMs have driven significant advancements across core AI domains, including natural language processing (NLP), computer vision (CV), as well as applications in medicine, law, education~\cite{bommasani2021opportunities, moor2023foundation, ray2023chatgpt}. Building on this success, multi-task learning (MTL) has been a paradigm for integrating task-specific abilities into a model~\cite{fifty2021efficiently}, allowing generalization across multiple specialized tasks. Nonetheless, MTL requires simultaneous training on all targeted datasets, which can be costly and pose privacy concerns. Model merging~\cite{Wortsman2022ModelSA,ilharco2022editing,du2024parameter} has recently emerged as an alternative paradigm to MTL for task adaptation, enabling the training-free integration of fine-tuned models, which are increasingly being shared publicly (e.g., on Hugging Face).

Model merging began with weight interpolation~\cite{Wortsman2022ModelSA} to combine the strengths of different models in weight space by balancing competition and cooperation within shared representation~\cite{du2024parameter,ilharco2022editing,ortiz2024task}. In NLP, methods such as merging task-specific language models have been explored to build and update foundation models with multi-task capabilities \cite{wan2024knowledge, akiba2025evolutionary, wan2025fusechat}. Similarly, in CV, approaches like merging Vision Transformers (ViTs) trained on different tasks or domains have been investigated to create unified models capable of handling diverse visual tasks \cite{kim2021vilt, bao2022vlmo, merge-vision}. In the multi-modal space, model merging has been applied to integrate models from different modalities, such as text and images, enhancing tasks like audio-visual question answering and image captioning~\cite{sung2023empirical, sundar2024multimodal, dziadzio2024merge}. These advancements underscore model merging as a promising avenue for future research.

\begin{figure}[t]
  \includegraphics[width=\columnwidth]{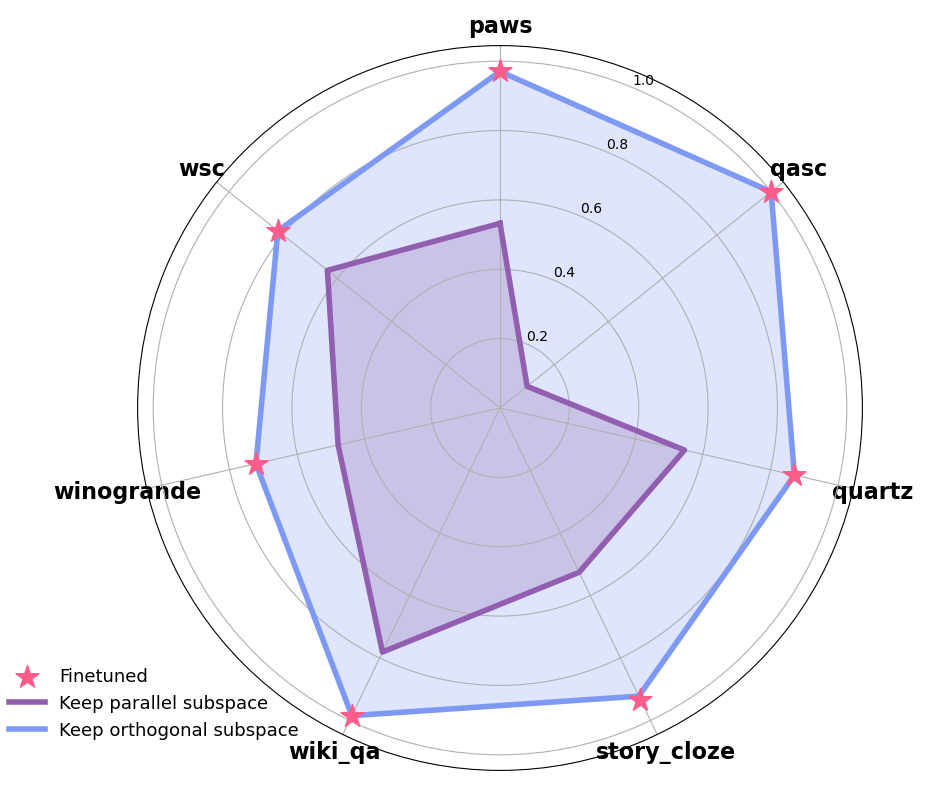}
  \caption{Impacts on neuronal subspaces decomposition of T5-Large. Retaining the orthogonal subspace while removing the parallel subspace preserves near-perfect performance across all tasks. In contrast, keeping the parallel subspace while removing the orthogonal subspace leads to a significant performance drop.}
  \label{fig:intro-exp}
\end{figure}

Existing methods for model merging primarily operate at three granularities: model-level, layer-level, or parameter-level~\cite{ilharco2022editing, yang2023adamerging, du2024parameter}, while overlooking the fundamental roles of neurons, their activation and connectivity~\cite{suhaimi2022representation,stelzer2021deep}, which underpins the learning process all neural networks from Perception~\cite{rosenblatt1958perceptron} to LLMs~\cite{touvron2023llama}. Table~\ref{tab:diff-granularity} provided a summary of existing methods across different granularities. In Figure~\ref{fig:intro-exp} and 
Table~\ref{tab:t5-large-intro-exp-discussion}
, we analyze on how changes in model weights along two complementary neuronal subspaces, leads to distinct impacts on task performance,  by removing one and retaining the other. Notably, one subspace preserves most of the task-specific capabilities. This observation motivates us to explore model merging at the neuronal level, which could have important implications for mitigating task interference and could yield more robust merged~models.

\begin{figure*}[!t]
  \centering
  \begin{overpic}[width=\linewidth, trim=53.2 80.3 52.8 42, clip]{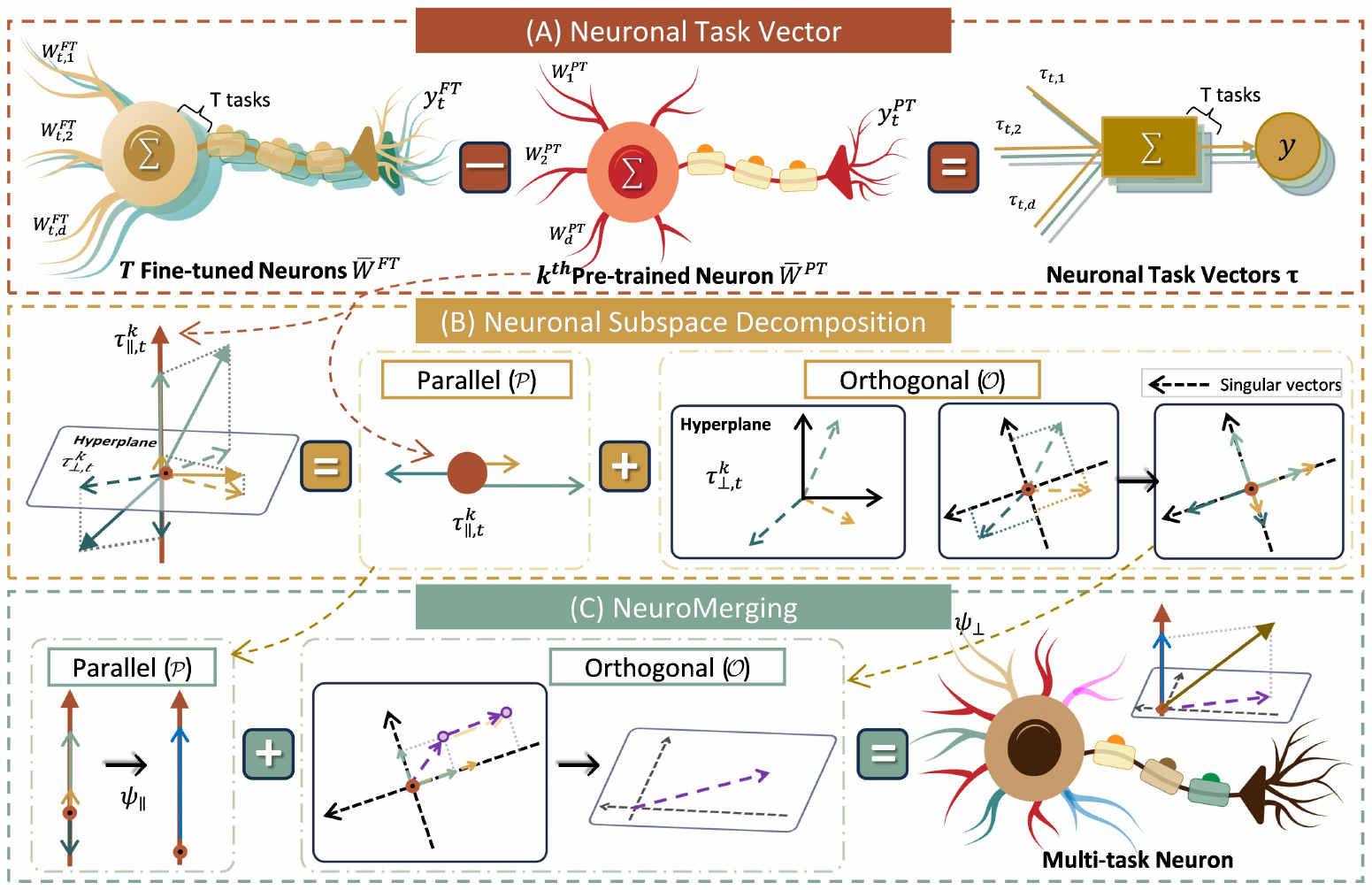}
    \put(15.7,48.2){\fontsize{8}{7.8}\selectfont \citep{rosenblatt1958perceptron}}
  \end{overpic}
  \caption {Illustration of our proposed framework. \textbf{(A)} Our approach explicitly considers neuronal activation mechanisms through \textit{neuronal task vectors} $\tau_t^k \in R^d$, defined as the difference between fine-tuned and pre-trained neurons for task $t$. \textbf{(B)} These vectors are decomposed into parallel and orthogonal subspaces relative to pre-trained neurons, corresponding to input sensitivity and task adaptability. SVD constructs a coordinate system in the previously unstructured orthogonal subspace. \textbf{(C)} Our \textit{\textbf{NeuroMerging}} method merges models efficiently within these low-dimensional complementary subspaces, rather than in the original high-dimensional weight space.}
  \label{fig:main}
\end{figure*}

In this work, we present the first study to examine model merging at the neuronal level, illustrated in Figure~\ref{fig:main}. Specifically, we investigate the roles of neuronal mechanisms (i.e., neuron connectivity and activation) in model merging, both mathematically and empirically. We begin by decomposing task-specific representations from fine-tuned models into two complementary neuronal subspaces. These subspaces are mathematically characterized to show that they regulate  neuron sensitivity and input adaptability. The key differences of our neuronal subspaces with existing task vectors summarized in Table~\ref{tab:task_vector_comparison}. Leveraging the insights from the decomposition, we introduce \textbf{NeuroMerging}, a novel merging framework designed to mitigate task interference within neuronal subspaces, enabling training-free model fusion across diverse tasks. To evaluate our approach, we conduct experiments on multi-task benchmarks across both natural language and vision domains, considering various settings, including in-domain and out-of-domain generalization. Empirically, our method outperforms existing approaches as summarized in Figure~\ref{fig:intro-sun}. The main contributions of our paper are as follows:

\begin{itemize}
    \item We presented the first exploration into the roles of neuronal mechanisms in the merging process, introducing a decomposition of task-specific representations into two complementary neuronal subspaces.
    \item Based on the insights from the neuronal subspaces, we proposed NeuroMerging, a new framework developed for model merging accounting input sensitivity and task specificity.
    \item We showed that NeuroMerging achieved superior performance compared to existing approaches on multi-task benchmarks in both natural language processing and vision domains.
\end{itemize}

%% file: tables/TVvsNTV.tex
\begin{table*}[t!]
\centering
\renewcommand{\arraystretch}{1.2}
\small
\begin{tabular}{>{\centering\arraybackslash}p{2cm}|
                >{\centering\arraybackslash}p{5.5cm}|
                >{\centering\arraybackslash}p{6cm}}
\thickhline
\rowcolor{mygray}
\textbf{Aspect} & \textbf{Task Vector} & \textbf{Neuronal Task Vector} \\
\hline
\textbf{Vector Space} & Resides in the weight space & Resides in two complementary neuronal subspaces of neurons \\
\textbf{Interpretation} & Represents weight interpolation between pre-trained and fine-tuned models & Aligns with neuronal mechanisms, with input sensitivity and task adaptability \\
\textbf{Activation} & $y = \phi\left(\left( \mathbf{w}_0^k + \tau_t^k \right) \cdot \bar{\mathbf{x}}\right)$ & 
$y = \phi\left(\mathbf{w}_0^k \cdot \bar{\mathbf{x}}_{\parallel} + \tau_{\parallel,t}^k \cdot \bar{\mathbf{x}}_{\parallel} + \tau_{\perp,t}^k \cdot \bar{\mathbf{x}}_{\perp}\right)$ \\
\thickhline
\end{tabular}
\caption{\label{tab:task_vector_comparison}Differences between task vector and neuronal task vector. Please refer to Section~\ref{sec:method} for details.}
\end{table*}

%% file: tables/intro_compare.tex
\begin{table}[t]
\centering
\resizebox{\columnwidth}{!}{
\begin{tabular}{r|c|c}
\thickhline
\rowcolor{mygray}
\textbf{Method} & \textbf{Scale} & \textbf{Granularity Level} \\ 
\hline
Fisher Merging\pub{NeurIPS22}     & Fisher Matrix  & Parameter \\
RegMean\pub{ICLR23}    & Inner Product Matrix  & Parameter \\
Task Arithmetic\pub{ICLR23}         & Uniformed  & Task \\
Ties-Merging\pub{NeurIPS23}       & Uniformed  & Parameter \\
DARE\pub{ICML24}       & $1/(1 - p)$  & Parameter \\
LoraHub\pub{COLM24}    & Evolver Searched  & Task \\
AdaMerging\pub{ICLR24} & Unsupervised Optimized  & Layer \\
PCB-Merging\pub{NeurIPS24}        & Balancing Matrix  & Parameter \\
\hline
\textbf{NeuroMerging (Ours)} & \textbf{L1-Norm} & \textbf{Neuron} \\
\thickhline
\end{tabular}
}
\caption{\label{tab:diff-granularity} Different merging scales and granularity levels.}
\end{table}

%% file: sections/2_RelatedWork.tex
\section{Related Work}
\label{Related Work}

Multi-Task Learning (MTL)~\cite{fifty2021efficiently} leverages transferable knowledge to handle multiple related tasks simultaneously. Existing MTL approaches primarily rely on architectural design or optimization strategies. Architectural-based methods, such as Mixture of Experts (MoE) \cite{shazeer2017outrageously}, introduce specialized subnetworks that dynamically route inputs to task-specific experts, effectively reducing interference. However, these methods require modifying the pretrained model structure, increasing computational complexity, and limiting scalability \cite{liu2019end, shen2024efficient, lu2024twin}. Optimization-based approaches, on the other hand, focus on balancing task gradients or loss functions to mitigate task conflicts during training \cite{bai2023qwen, kendall2018multi}. While these methods improve convergence, they still depend on task-specific training data, which may be impractical in real-world applications due to privacy concerns or data scarcity \cite{liang2020model}. In contrast, model merging offers an alternative paradigm by integrating knowledge from multiple fine-tuned models into a single unified model without requiring additional training data or architectural modifications~\cite{Wortsman2022ModelSA,ilharco2022editing}. Notwithstanding the promising findings, a key challenge in model merging is task conflict~\cite{yadav2024ties,du2024parameter}, where different tasks compete for model capacity, potentially leading to suboptimal performance.

To resolve task conflicts, existing model merging methods can be categorized into three levels based on their granularity. Model-level merging combines entire model weights, typically through averaging or weighted aggregation, but often results in performance degradation due to the loss of task-specific knowledge, as seen in methods like Task Arithmetic \cite{ilharco2022editing} and LoRAHub \cite{huang2023lorahub}. Layer-level merging selectively integrates layers from different models under the assumption of shared representations; for instance, AdaMerging \cite{yang2023adamerging} adapts layer selection to better preserve task-specific information. Parameter-level merging directly manipulates individual parameters to blend knowledge from multiple models, enhancing adaptability and robustness. Techniques such as Fisher Merging \cite{matena2022merging}, RegMean \cite{jin2022dataless}, TIES-Merging \cite{yadav2024ties}, DARE \cite{yu2024language}, and PCB-Merging \cite{du2024parameter} exemplify this approach. However, existing methods largely overlook the fundamental role of neuron and neuron-level interactions in task specialization. In this work, we aim to bridge this~gap. 

Current neural network models, from the Perceptron invented by~\citet{rosenblatt1958perceptron} to recent massive LLMs~\cite{touvron2023llama}, have grown significantly in scale and complexity. Nevertheless, the core principle remains unchanged: neurons and their connectivity still underpin the learning process~\cite{stelzer2021deep,suhaimi2022representation}. During the pre-training and fine-tuning process, neurons are not merely passive components but active elements of the network, each contributing to learning and inference~\cite{jiang2024network,islam2023revealing}. In this work, we attempt to conduct the first in-depth study on neuronal mechanisms in model merging.

%% file: sections/3_Methodology.tex
\section{Methodology}\label{sec:method}

In this section, we first formalize the concept of neuronal task vector for model merging and then decompose neuronal task vectors into two complementary neuronal subspaces. Subsequently, 
we introduce our framework, \textbf{NeuroMerging}, shown in Algorithm~\ref{algo} in Appendix~\ref{sec:algo}, which performs merging in the neuronal subspaces. 

\subsection{Preliminaries}\label{premethod}
Model merging considers the fusion of a set of \( T \) task specific models, \( (\theta_{1}, \dots, \theta_{t}, \dots, \theta_{T}) \), fine-tuned from a pretrained model \( \theta_{0} \). With the task vector notation, each task is defined as \( \tau_t = \theta_{t} - \theta_{0} \). The merged model is $\bar{\theta} = \theta_{0} + \xi (\tau_1,...,\tau_t,...,\tau_T)$, where $\xi(\cdot)$ represents the transformation applied to each task vector \( \tau_t \) and followed by merging.

Zooming into the neuronal level of \( \tau_t \) from any neural network models (from perceptron to LLMs), the activation of the neuron can be computed with an input $\bar{x} \in \mathbb{R}^n$:
\begin{equation}
    y = \phi(\mathbf{w}^k_t \cdot \bar{x})
\end{equation}
\noindent where $\mathbf{w}^k_t$ represents the synaptic connectivity to the $k^{th}$ neuron after fine-tuning from the pre-trained weights $\mathbf{w}^k_0$. $\phi(\cdot)$ denotes non-linear activation function. Noting that although the input $\bar{x}$ is involved in the definition and derivation, it is not required during the merging process.

\subsection{Neuronal Task Vector}\label{NTA}
Neuronal task vector is defined to be the difference between task-specific fine-tuning of a neuron with the pre-trained model: $\mathbf{\tau}^k_t = \mathbf{w}^k_t - \mathbf{w}^k_0$.
Based on the defined neuronal task vector, we can rewrite the activation of the $k^{th}$ neuron as:

\begin{equation}
    y = \phi((\mathbf{w}^k_0 + \tau^k_t) \cdot \bar{x})
\end{equation}

\subsection{Neuronal Subspace Decomposition}

To examine how task-specific fine-tuning impact neurons, we decompose the neuronal task vectors into two complementary neuronal subspaces, visualized in Figure~\ref{fig:main}. Mathematically, this decomposition is formulated as:
\begin{equation}
     \tau^k_t = \tau^k_{\parallel, t} + \tau^k_{\perp, t},
\end{equation}

\noindent where $\tau^k_{\parallel, t} = \mathbf{P} \tau^k_t$ projects the neuronal task vectors onto the pre-trained model's weight space, \textbf{Parallel Subspace} ($\mathcal{P}$).
Here, $\mathbf{P}$ is the projection matrix onto the span of $\mathbf{W}^k_0$. $\tau^k_{\perp, t} = (\mathbf{I} - \mathbf{P}) \tau^k_t$ captures the complementary orthogonal modifications, in the \textbf{Orthogonal Subspace} ($\mathcal{O}$). Noting that $\bar{x}$ is not required during decomposition.

Based on Neuronal Subspace Decomposition, we decompose both $\tau^k_t$ and $\bar{x}$ into two orthogonal subspaces ($\mathbf{P}$ and $\mathcal{O}$) of $\mathbf{w}^k_0$ and obtain:

\begin{equation}
    y = \phi((\mathbf{w}^k_0 + \tau^k_{\parallel, t} + \tau^k_{\perp, t}) \cdot (\bar{x}_{\parallel} + \bar{x}_{\perp}))
\end{equation}

Due to the orthogonality between vectors in the parallel subspace and those in the orthogonal subspace, we can simplify the expression as:

\begin{equation}
    y = \phi(\mathbf{w}^k_0 \cdot \bar{x}_{\parallel} + \tau^k_{\parallel, t} \cdot \bar{x}_{\parallel} + \tau^k_{\perp, t} \cdot \bar{x}_{\perp})
\end{equation}

Since $\mathbf{w}^k_0$ and $\tau^k_{\parallel, t}$ are parallel, we can further express as:

\begin{equation}
    y = \phi(s(\tau^k_{\parallel, t}) \cdot \bar{x}_{\parallel} + \tau^k_{\perp, t} \cdot \bar{x}_{\perp})
\end{equation}

Thus, we obtain a mechanistic view of the fine-tuned neuron: it adjusts its sensitivity $s$ to input in the parallel subspace ($\mathcal{P}$) while adapting itself to capture $\bar{x}_{\perp}$ to handle task $t$ in the orthogonal subspace ($\mathcal{O}$).

The role of each complementary subspace:
\begin{itemize}
    \item \textbf{Parallel Subspace} ($\mathcal{P}$): this subspace captures transformations that preserve shared representations with the $\mathbf{w}^k_0$. It is also closely related to neuron sensitivity with larger magnitudes corresponding to higher sensitivity to changes in input activations.
    \item \textbf{Orthogonal Subspace} ($\mathcal{O}$): The orthogonal complementary subspace of $\mathcal{P}$ represents novel task-specific adaptations introduced during fine-tuning, capturing input adaptability to task specific representation.
\end{itemize}

\subsection{NeuroMerging}\label{NeuroMerging}
Our proposed NeuroMerging merges neuronal task vectors along two complementary neuronal spaces: 
\begin{align}
\overline{\tau}^k &= \lambda_1\psi_\parallel \left( \tau^k_{\parallel, 1}, \dots, \tau^k_{\parallel, t}, \dots, \tau^k_{\parallel, T} \right) \\
&\quad + \lambda_2\psi_{\perp} \left( \tau^k_{\perp, 1}, \dots, \tau^k_{\perp, t}, \dots, \tau^k_{\perp, T} \right) \nonumber
\end{align}

\noindent where $\psi_\parallel(\cdot)$ denotes the merging function for $\tau^k_{\parallel, t} \in \mathbb{R}$, which can be a commonly used weighted average, TIES's disjoint merge~\cite{yadav2024ties}, and others. $\psi_\parallel(\cdot)$ is also applied to non-neuronal parameters such as bias and pre-norm. For $\tau^k_{\perp, t} \in \mathbb{R}^{T}$, we first find dominant orthogonal subspaces within $\mathcal{O}$ using singular value decomposition (SVD) with rank being the number of tasks interact within the same neuron, and then project $\tau^k_{\perp, t}$ along each dimension of the SVD subspace, before applying $\psi_\parallel(\cdot)$ to them. Detailed discussion on SVD is provided in Appendix \ref{sec:SVD}.

$\lambda_1$ and $\lambda_2$ are scaling parameters. When validation data is available $\lambda_1$ and $\lambda_2$ could be tuned. However, when validation is unavailable, we propose setting $\lambda_1 = 0$ (as the corresponding subspace is observed to have little impact in Section~\ref{Robustlambda}) and estimate $\lambda_2$ based on the impact of top \( r\% \) mentioned in Section~\ref{premethod}. Specifically, $\lambda_2$ is estimated as $\lambda_2 = \frac{1}{1 - \sigma}$, where \( \sigma = max(\sigma_1,...,\sigma_t,...,\sigma_T)\) and $\sigma_t = \frac{\|\mathbf{\tau}_t^{masked}\|_1 }{\|\mathbf{\tau}_t\|_1}$ is the ratio of the $L_1-Norm$ of the zeroed-out elements in the task vector $ \mathbf{\tau}_t^{masked} $ to the $L_1-Norm$ of the original task vector $ \mathbf{\tau}_t $. Detailed discussion on the parameters is provided in Section~\ref{Robustlambda}.

Subsequently, we reconstruct the task vector \( \overline{\tau}\) with the all the merged neuronal task vectors \( \overline{\tau}_k\). Finally, we obtained the final merged model $\overline{\theta} = \theta_0 + \overline{\tau}_{rescaled}$. The complete procedure is presented in Algorithm~\ref{algo}, with additional discussion on the motivation provided in Appendix~\ref{sec:Statistical-Analysis-of-Parameters}.

%% file: sections/4_ExperimentalSetup.tex
\section{Experimental Setup}

\input{tables/T5-Large}

\textbf{Baseline Methods.} 
Our baselines comprise two main categories: (1) non-model merging approaches, which include individually fine-tuned models and a multitask model trained jointly on the combined dataset serving as our theoretical upper bound, and (2) various advanced model merging techniques, including 
Simple Averaging \citep{Wortsman2022ModelSA}, 
Fisher Merging \citep{matena2022merging}, 
RegMean \citep{jin2022dataless}, 
Task Arithmetic \citep{ilharco2022editing}, 
TIES-Merging \citep{yadav2024ties}, 
AdaMerging \mbox{\citep{yang2023adamerging}}, 
PCB-Merging \citep{du2024parameter}, 
NPS \citep{du-etal-2025-neural}, 
and our proposed NeuroMerging method. 
Notably, we used task-wise AdaMerging for fair comparison with existing baselines and our method. 
See Section~\ref{sec:layer_wise_merging} for further discussion and analysis.
We reported average accuracy across all tasks' test sets as our primary evaluation metric.

\noindent \textbf{Validation Set Availability.}
Previous works exhibited varying dependencies on a validation set. Fisher Merging inherently required a validation set to compute the Fisher matrix. Other approaches may optionally utilize validation data for hyperparameter tuning, while RegMean leverages training data to compute and store inner product matrices for model merging. However, since these matrices matched the dimensions of the original model, they introduced substantial storage and computational overhead, limiting scalability to larger models and more extensive merging tasks. 

Task vector-based approaches such as Task Arithmetic, Ties-Merging, and PCB-Merging, along with our proposed NeuroMerging, are substantially more lightweight and efficient. These approaches are training-free and do not rely on a validation set, making them highly practical for real-world applications. To further evaluate this advantage, we conducted additional experiments comparing task vector-based methods in scenarios where validation sets were unavailable.

\noindent \textbf{Hyperparameters.}
In the absence of an additional validation set, we set $ \lambda = 1 $ as the default value for all task-vector-based methods. For TIES-Merging and PCB-Merging, which required a masking ratio, we followed the settings of \citet{yadav2024ties} and \citet{du2024parameter}, applying $ r = 0.2 $ as the default value across all experiments. For NeuroMerging, we set a default masking ratio of $ r = 0.15 $, with $ \lambda_1 $ fixed at $0$, while $ \lambda_2 $ was automatically adjusted according to the methodology described in Section~\ref{NeuroMerging}. 

When validation is allowed, we configure the non-diagonal multiplier $ \alpha $ in RegMean to $ 0.9 $, except for the T5-base model, where it is set to $ 0.1 $. For Task Arithmetic, we performed a grid search over $ \lambda $ ranging from $ 0.2 $ to $ 1.5 $ with a step size of $ 0.1 $. 
For TIES-Merging, PCB-Merging, and NeuroMerging, we searched for the optimal masking ratio r in the range $ [0.05,0.2] $ with a step size of $ 0.05 $, and $ \lambda $ ($ \lambda_2 $ for NeuroMerging) from $ 0.8 $ to $ 5.0 $ with a step size of $ 0.1 $. 

%% file: tables/T5-Large.tex
\begin{table*}[t]
\centering
\captionsetup{type=table}
\resizebox{0.9\linewidth}{!}{  
\begin{tabular}{r|c|c|ccccccc}
\thickhline
\rowcolor{mygray}\textbf{Task($\rightarrow$)} &  &   & \multicolumn{7}{c}{\textbf{Test Set Performance}} \\ 
\cline{4-10}
\rowcolor{mygray}\textbf{Method($\downarrow$)} & \multirow{-2}{*}{\textbf{Validation}} & \multirow{-2}{*}{\textbf{Average}}  & paws & qasc & quartz & story\_cloze & wiki\_qa & winogrande & wsc \\ 
\hline
Zeroshot  & -  & 53.1  & 58.2  & 54.2  & 54.1  & 54.3  & 70.9  & 49.2  & 63.9 \\
Finetuned & -  & 88.0  & 94.4  & 97.1  & 85.3  & 91.0  & 95.7  & 71.6  & 80.6 \\
Multitask & -  & 88.1  & 94.2  & 98.5  & 89.3  & 92.0  & 95.4  & 73.5  & 73.6 \\
\hline
Averaging\pub{ICML22} & \ding{55}  & 51.6  & 59.2  & 26.3  & 69.6  & 53.8  & 67.3  & 49.1  & 36.1 \\
Task Arithmetic\pub{ICLR23}        & \ding{55}  & 59.7  & 60.9  & 31.7  & 57.8  & 73.0  & 73.5  & 55.7  & \underline{65.3} \\
TIES-Merging\pub{NeurIPS23}      & \ding{55}  & 76.7  & 80.8  & 92.4  & 77.7  & 81.9  & \underline{78.4}  & \underline{61.9}  & 63.9 \\
PCB-Merging\pub{NeurIPS24}       & \ding{55}  & \underline{76.9}  & \textbf{82.9}  & \underline{93.2}  & \underline{79.0}  & \underline{84.4}  & 75.6  & \textbf{63.5}  & 59.7 \\
\textbf{NeuroMerging (Ours)} & \ding{55}  & \textbf{77.6}  & \underline{81.1}  & \textbf{94.3}  & \textbf{81.6}  & \textbf{84.7}  & \textbf{81.2}  & 56.7  & \textbf{83.9} \\
\hline
Fisher Merging\pub{NeurIPS22}   & \checkmark  & 68.7  & 68.4  & 83.0  & 65.5  & 62.4  & \textbf{94.1}  & 58.2  & 49.2 \\
RegMean\pub{ICLR23}  & \checkmark  & 79.8  & 83.9  & \textbf{97.2}  & 73.2  & 82.6  & \textbf{94.1}  & 63.2  & 64.4 \\
Task Arithmetic\pub{ICLR23}       & \checkmark  & 74.6  & 72.7  & 91.3  & \textbf{76.4}  & 85.6  & 74.4  & 61.0  & 61.1 \\
TIES-Merging\pub{NeurIPS23}     & \checkmark  & 79.5  & 82.6  & 94.9  & 72.8  & 87.4  & 85.2  & \textbf{66.6}  & 66.7 \\
PCB-Merging\pub{NeurIPS24}      & \checkmark  & \underline{81.0}  & \textbf{87.0}  & \underline{95.2}  & \textbf{76.4}  & \textbf{88.1}  & 88.4  & 64.3  & \underline{68.1} \\
\textbf{NeuroMerging (Ours)}  & \checkmark  & \textbf{82.2}  & \underline{86.4}  & 94.3  & \underline{75.9}  & \underline{87.9}  & \underline{91.2}  & \underline{65.9}  & \textbf{73.6} \\
\thickhline
\end{tabular}
}
\caption{\label{tab:t5-large} Test set performance when merging T5-Large models on seven NLP tasks.}
\end{table*}

%% file: sections/5_Results.tex
\section{Results}\label{sec:Results}

\subsection{Merging NLP Models}\label{sec:Merging-NLP-Models}
Following the experimental settings from \citet{yadav2024ties}, we used the T5-base and T5-large models \citep{raffel2020exploring}, which were encoder-decoder transformers \citep{vaswani2017attention} pretrained via masked language modeling on a large text corpus, and fine-tuned them independently on seven tasks: 
\citeposs{Khot2019QASC} \mbox{QASC}, 
\citeposs{yang-etal-2015-wikiqa} \mbox{WikiQA}, 
and \citeposs{tafjord-etal-2019-quartz} \mbox{QuaRTz} for Question Answering; 
\citeposs{paws2019naacl} \mbox{PAWS} for Paraphrase Identification; 
\citeposs{sharma-etal-2018-tackling} Story Cloze for Sentence Completion; 
and \citeposs{sakaguchi2021winogrande} \mbox{Winogrande} together with \citeposs{levesque2012winograd} \mbox{WSC} for Coreference Resolution. 
Tables \ref{tab:t5-large} and \ref{tab:t5-base} demonstrated that our approach achieved superior performance over state-of-the-art methods, achieving improvements of $ 0.6\% $ and $ 1.2\% $ for T5-base and T5-large, respectively. Moreover, NeuroMerging without validation showed even more substantial gains, surpassing previous methods by $ 1.4\% $ for T5-base and $ 0.7\% $ for T5-large. For comprehensive results across all tasks and model variants, see Appendix~\ref{sec:Comprehensive-Task-Level-Results} Tables~\ref{tab:t5-large} and \ref{tab:t5-base}, with detailed error analysis provided in Appendix~\ref{error_analysis}.

\subsection{Out-of-Domain Generalization}\label{sec:Out-of-Domain-Generalization}
\begin{figure}[h!]
  \includegraphics[width=0.95\columnwidth]{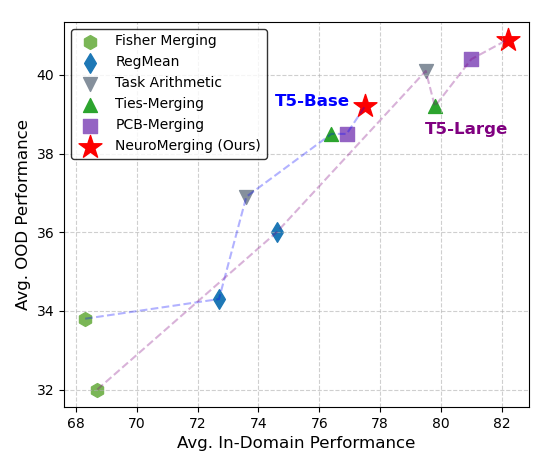}
  \caption{In-domain v.s. Out-domain performance.}
  \label{fig:in-out-domain}
\end{figure}

Building upon the experimental setup of \citet{yadav2024ties}, we also investigated how merged models between tasks enhance generalization in different domains. 
Following the approach used in prior NLP models, we merged models on seven in-domain datasets and evaluated their performance on six held-out datasets from the T0 mixture \citep{sanh2022multitask} to assess cross-task generalization.
These datasets encompass diverse tasks, covering three Question Answering datasets: 
\citeposs{huang-etal-2019-cosmos} \mbox{Cosmos QA}, \citeposs{sap2019social} \mbox{Social IQA}, and \citeposs{Rogers_Kovaleva_Downey_Rumshisky_2020} \mbox{QuAIL}; 
one Word Sense Disambiguation dataset: \citeposs{pilehvar-camacho-collados-2019-wic} \mbox{WiC}; 
and two Sentence Completion datasets: \citeposs{gordon-etal-2012-semeval} \mbox{COPA} and \citeposs{zellers-etal-2019-hellaswag} \mbox{H-SWAG}. 
As shown in Figure \ref{fig:in-out-domain}, NeuroMerging outperformed the strongest baseline by $ 0.7\% $ and $ 0.5\% $ for T5-base and T5-large models, respectively, showcasing enhanced out-of-domain generalization capabilities. For more comprehensive results, please refer to Appendix \ref{sec:Comprehensive-Task-Level-Results} Tables \ref{tab:t5-base-ood} and \ref{tab:t5-large-ood}, with detailed error analysis provided in Appendix~\ref{error_analysis}.

\subsection{Merging LLMs}\label{sec:Merging-LLMs}
\input{tables/LLaMA2}
We followed the experimental setup of \citet{du2024parameter} and extended our NeuroMerging to larger LLMs. 
Specifically, we merged three domain-specialized Llama-2-7b models \citep{touvron2023llama}, and each was fine-tuned for distinct capabilities: 
Chinese language understanding, 
mathematical reasoning \citep{yu2024metamath}, 
and code generation \citep{roziere2023code}. 
To rigorously evaluate the performance of each specialized model, we employed established benchmarks tailored to their respective domains: 
\citeposs{li-etal-2024-cmmlu} CMMLU for assessing Chinese language proficiency, 
\citeposs{cobbe2021training} GSM8K for mathematical capabilities, 
and \citeposs{chen2021evaluating} HumanEval for code generation competency. 
As demonstrated in Table \ref{tab:llama2}, our approach exhibited substantial performance improvements, surpassing the strongest baseline by $ 0.9\% $. Notably, our method even outperformed the PCB-Merging utilizing Evolution Strategies (ES) optimization algorithm by $ 0.7\% $, underscoring the effectiveness of our proposed methodology. Appendix~\ref{error_analysis} provides the error analysis.

\subsection{Merging Vision Models}\label{sec:Merging-Vision-Models}
\input{tables/ViT_base_large_small}
We also examined the modality of vision by adhering to the experimental setup outlined by \citet{ilharco2022patching, ilharco2022editing}. 
Specifically, we adopted two variants of the CLIP model \citep{radford2021learning}, ViT-B/32 and ViT-L/14, as visual encoders \citep{dosovitskiy2021an}. 
For a fair comparison, we obtained fine-tuned checkpoints from \citet{ilharco2022editing}, which were consistently utilized across all baseline methods. 
This comprehensive evaluation spans multiple classification domains, encompassing remote sensing, traffic analysis, and satellite imagery recognition, with evaluations conducted on standard benchmark datasets, including 
\mbox{Cars} \citep{2013Cars}, 
\mbox{DTD} \citep{2014DTD}, 
\mbox{EuroSAT} \citep{2019EuroSAT}, 
\mbox{GTSRB} \citep{2011GTSRB}, 
\mbox{MNIST} \citep{lecun1998mnist}, 
\mbox{RESISC45} \citep{2017RESISC45}, 
\mbox{SUN397} \citep{2016SUN397}, 
and \mbox{SVHN} \citep{2011SVHN}. 
Table~\ref{tab:vit-b-l} presented the results of NeuroMerging, demonstrating its competitive performance across different validation scenarios. When employing validation data, our method achieved performance improvements of $ 0.2\% $ for ViT-B/32 and $ 0.8\% $ for ViT-L/14 over state-of-the-art baselines. In the absence of additional validation, NeuroMerging further improved upon the strongest baseline by $ 0.5\% $ and $ 1.0\% $ for ViT-B/32 and ViT-L/14, respectively. These results substantiated the broad model compatibility of our approach. For comprehensive results across all tasks and model variants, see Appendix~\ref{sec:Comprehensive-Task-Level-Results} Tables~\ref{tab:vit-b-32} and \ref{tab:vit-l-14}, with detailed error analysis provided in Appendix~\ref{error_analysis}.

\input{tables/ablation_intro}

%% file: tables/LLaMA2.tex
\begin{table}[ht]
\centering
\captionsetup{type=table}
\resizebox{\linewidth}{!}{  
\begin{tabular}{r|ccc|c}
\thickhline
\rowcolor{mygray}
Model & CMMLU & GSM8K & Human-Eval & Average \\ 
\hline
Chinese  & 38.6  & 2.3  & 13.4  & 18.1 \\
Math     & 31.2  & 65.6  & 0.0  & 32.3 \\
Code     & 33.3  & 0.0  & 17.1  & 16.8 \\
\hline
Averaging\pub{ICML22} & 35.6  & 48.5  & 6.7  & 30.3 \\
Task Arithmetic\pub{ICLR23}       & 35.4  & 46.1  & 9.8  & 30.4 \\
TIES-Merging\pub{NeurIPS23}     & \textbf{36.5}  & 53.4  & 12.8  & 34.3 \\
Consensus TA\pub{ICML24} & - & - & - & 33.5 \\
Consensus TIES\pub{ICML24} & - & - & - & 34.4 \\
PCB-Merging\pub{NeurIPS24}      & 36.4  & 52.3  & \textbf{16.5}  & 35.1 \\
PCB-Merging+ES\pub{NeurIPS24}   & 36.4  & 53.1  & \textbf{16.5}  & \underline{35.3} \\
NPS\pub{ACL25} & - & - & - & \underline{35.3} \\
\textbf{NeuroMerging (Ours)} & 36.1  & \textbf{57.2}  & 14.6  & \textbf{36.0} \\
\thickhline
\end{tabular}
}
\caption{\label{tab:llama2} Performance comparison on LLaMA2.}
\end{table}

%% file: tables/ViT_base_large_small.tex
\begin{table}[b!]
\centering
\captionsetup{type=table}
\resizebox{\linewidth}{!}{  
\begin{tabular}{r|c|c|c}
\thickhline
\rowcolor{mygray}
\textbf{Method} & \textbf{Validation} & \textbf{ViT-B/32 Avg.} & \textbf{ViT-L/14 Avg.} \\ 
\hline
Individual    & -  & 90.5  & 94.2  \\
Multi-task    & -  & 88.9  & 93.5  \\
\hline
Averaging\pub{ICML22}     & \ding{55}  & 65.8  & 79.6  \\
Task Arithmetic\pub{ICLR23} & \ding{55}  & 60.4  & 83.3  \\
TIES-Merging\pub{NeurIPS23}  & \ding{55} & 72.4  & 86.0  \\
PCB-Merging\pub{NeurIPS24}   & \ding{55}  & \underline{75.9}  & \underline{86.9}  \\
\textbf{NeuroMerging (Ours)} & \ding{55}  & \textbf{76.4}  & \textbf{87.9}  \\
\hline
Fisher Merging\pub{NeurIPS22}        & \checkmark  & 68.3  & 82.2  \\
RegMean\pub{ICLR23}       & \checkmark  & 71.8  & 83.7  \\
Task Arithmetic\pub{ICLR23} & \checkmark & 70.1  & 84.5  \\
TIES-Merging\pub{NeurIPS23}  & \checkmark & 73.6  & 86.0  \\
AdaMerging\pub{ICLR2024}  & \checkmark & 71.1  & 84.9  \\
AdaMerging++\pub{ICLR2024}  & \checkmark & 73.7  & 87.3  \\
PCB-Merging\pub{NeurIPS24}   & \checkmark & \underline{76.3}  & \underline{87.5}  \\
\textbf{NeuroMerging (Ours)} & \checkmark & \textbf{76.5}  & \textbf{88.3}  \\
\thickhline
\end{tabular}
}
\caption{\label{tab:vit-b-l} Performance comparison on ViT.}
\end{table}

%% file: tables/ablation_intro.tex
\begin{table}[t]
\centering
\resizebox{\linewidth}{!}{  
\begin{tabular}{c|c c c}
\thickhline
\rowcolor{mygray}
\textbf{T5-large} & \textbf{In-domain} & \textbf{Out-domain} & \textbf{Total Average} \\ 
\hline
Fine-tuned      & 88.0  & 53.8  & 58.7  \\
\hline
Keep Orthogonal   & \textbf{88.0}  & \textbf{53.9}  & \textbf{58.8}  \\
Keep Parallel   & 52.7  & 51.8  & 51.9  \\
\thickhline
\end{tabular}
}
\caption{\label{tab:t5-large-intro-exp-discussion} Role of orthogonal and parallel subspaces on T5-Large, details provided in  Appendix Table~\ref{tab:t5-large-intro-exp-all-table}.}
\end{table}

%% file: sections/6_AdditionalResults.tex
\section{Additional Results and Analysis}
\subsection{Merging without Validation Sets}\label{sec:Merging-without-Validation-Sets}
When validation data is unavailable, we examined the parameters \( \lambda_1 \) and \( \lambda_2 \) selected according to Section~\ref{NeuroMerging}, where \( \lambda_1 \) was set to zero, as the corresponding subspace was observed to have little impact in Section~\ref{Robustlambda}. The value of \( \lambda_2 \) was computed based on the L1-norm of masked and unmasked task vectors. 
Figure~\ref{fig:t5-large-vit-large-non-val} and Table~\ref{tab:t5-large} and Appendix Tables~\ref{tab:t5-large}, ~\ref{tab:vit-b-32}, and~\ref{tab:vit-l-14} presented the evaluation of NeuroMerging on NLP and CV tasks across various model sizes, comparing it with existing methods. NeuroMerging achieves the highest average accuracy across all tasks. This demonstrated that our proposed method, with a simple rescaling, outperformed existing methods on average. Specifically, it achieved a $1.4\%$ and $0.7\%$ improvement over the strongest baseline for T5-Base and T5-Large, respectively. For vision models, it outperformed the strongest baseline by $0.5\%$ and $1.0\%$ for ViT-B/32 and ViT-L/14, respectively.

\subsection{Role of Neuronal Subspaces}\label{subspaceroles}
Figure~\ref{fig:intro-exp}, Table~\ref{tab:t5-large-intro-exp-discussion} and Appendix Table~\ref{tab:t5-large-intro-exp-all-table} illustrated the impacts of neuronal subspace decomposition on T5-Large. To examine the impact of each subspace, ablation was performed separately on each subspace by retaining one while removing the other. Retaining the orthogonal subspace while removing the parallel subspace preserved near-perfect performance of finetuned checkpoints or even improved them across most tasks for T5-Large, achieving $88.0\%$ in-domain, $53.9\%$ out-of-domain, and an average of $58.8\%$. In contrast, keeping the parallel subspace while removing the orthogonal subspace resulted in a significant performance drop.

\begin{figure}[b]
  \centering
  \includegraphics[width=\columnwidth, trim=5 15 5 45, clip]{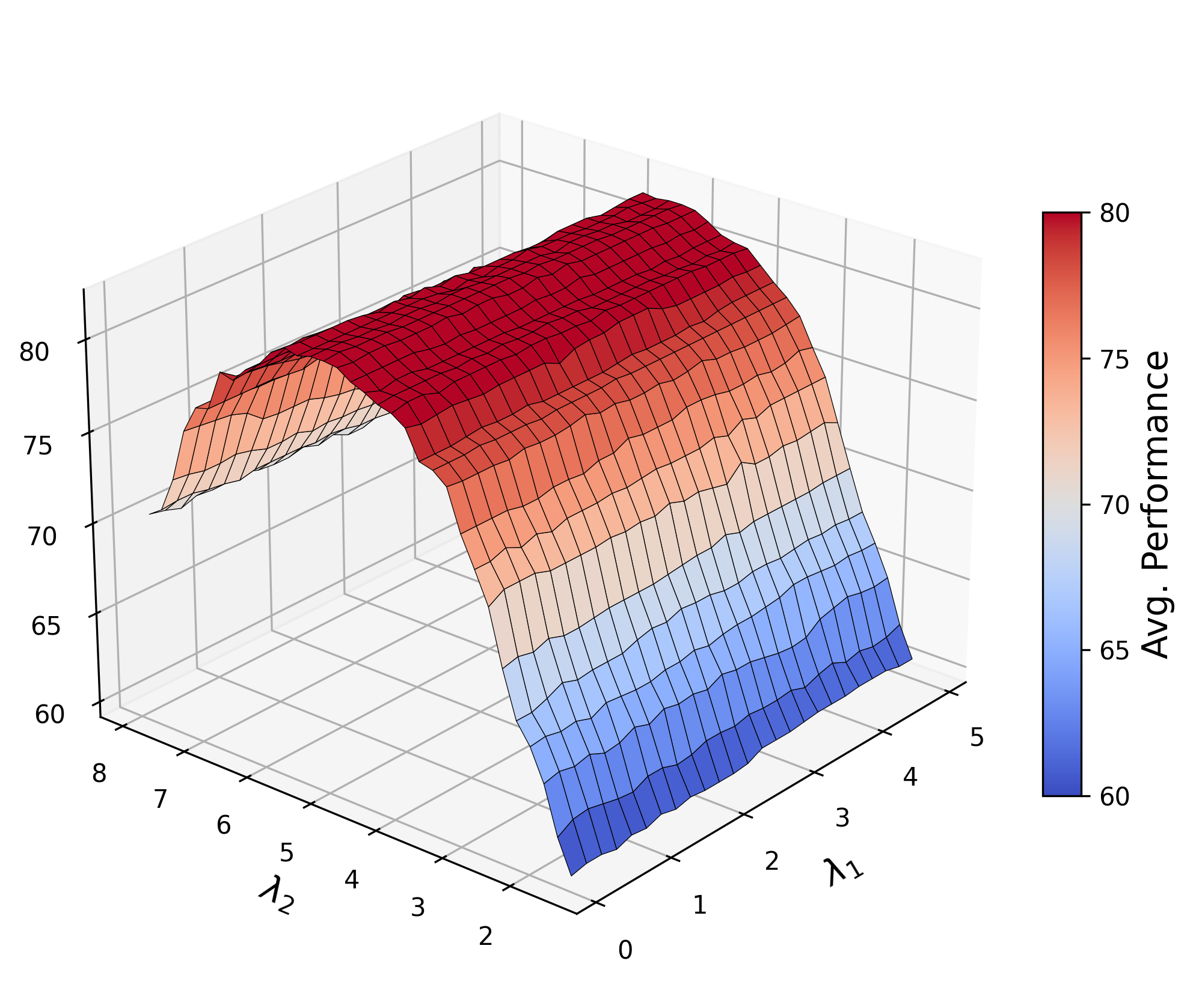}
  \caption{Impacts of $\lambda_1$ and $\lambda_2$ on T5-Large.}
  \label{fig:lam1-lam2}
\end{figure}

\subsection{Robustness of Hyperparameters}\label{Robustlambda}
We systematically investigated the impact of hyperparameters on merging performance: $ \lambda_1 $ and $ \lambda_2 $, which control the parallel and orthogonal subspace contributions, respectively, and the mask ratio $ r $.

\noindent \textbf{Relationship Between $ \lambda_1 $ and $ \lambda_2.$}
To examine the effects of $ \lambda_1 $ and $ \lambda_2 $, we conducted a grid search with $ \lambda_1 \in [0, 1.0] $ and $ \lambda_2 \in [3.0, 4.0] $ at $ 0.1 $ intervals, fixing $ r = 10\% $.
As visualized in Figure~\ref{fig:lam1-lam2}, the performance exhibited column-wise uniformity in the heatmap, indicating insensitivity to variations in $ \lambda_1 $, which aligned with our earlier discussion on the role of the orthogonal subspace in Section~\ref{subspaceroles}.
This highlights a practical strength of our approach: when computational resources are limited, users can simplify their merging tasks by primarily tuning $ \lambda_2 $, as adjusting $ \lambda_1 $ yields relatively subtle effects.
The optimal performance occurred at $ \lambda_2 = 3.6 $, attributed to the substantial proportion of masked variables.

\begin{figure}[h]  \includegraphics[width=\columnwidth]{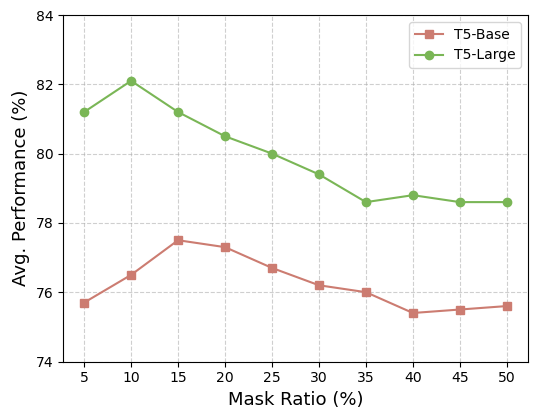}
  \caption{Impacts on mask ratio $r$.}
  \label{fig:ratio-vs-performance}
\end{figure}

\noindent \textbf{Masking Ratio $ r $.} 
When validation is available, we analyzed the impact of the mask ratio on performance, showed in Figure~\ref{fig:ratio-vs-performance}. 
We observed robust performance across different ratios, with accuracy peaked at $ 15\% $ and $ 10\% $ for T5-Base and T5-Large, respectively. Performance variations remained bounded (within $ 2.5\% $ for T5-Base and $ 4\% $ for T5-Large) and stabilized beyond $ 35\% $. Maintaining $r$ at $ 10\% \text{--} 20\% $ improved performance, whereas exceeding this range often led to degradation. 
This suggested that an optimal masking ratio balances information retention and redundancy reduction.

\subsection{Ablation Study on Merging Functions}\label{sec:merging_functions}
We conducted ablation experiments on various merging functions $\psi(\cdot)$ to evaluate their effectiveness in combining numerics. 
As shown in Table~\ref{tab:ablation}, among all merging functions, the \textit{elect+mean} approach from TIES-Merging achieves the highest performance at $ 82.2\% $. In comparison, using \textit{elect+sum}, \textit{averaging}, and \textit{sum} methods resulted in performance decreases of $ 2.6\% $, $ 2.5\% $, and $ 2.4\% $, respectively. 

\input{tables/ablation_phy}

\subsection{NeuroMerging + AdaMerging}\label{sec:layer_wise_merging}
We adopted task-wise AdaMerging~\citep{yang2023adamerging} for fair comparison, as the layer-wise variant incurs substantial computational cost due to tuning individual layer- and tasks-specific $\lambda$ values via expensive backpropagation on a model larger than the pre-trained one. Further computational efficiency analysis between different methods is detailed in Appendix~\ref{sec:scalability_eval}.

As shown in Table~\ref{tab:layer-wise-enhancement}, our method consistently improves the performance of both layer-wise AdaMerging and AdaMerging++ when applied to their publicly shared layer-wise $\lambda$ directly. 
Notably, we observe a greater improvement when combining AdaMerging with NeuroMerging compared to AdaMerging++, suggesting that the TIES component within AdaMerging++ may inadvertently discard information beneficial to NeuroMerging.

\input{tables/ablation_adamerging}

%% file: tables/ablation_phy.tex
\begin{table}[ht]
\centering
\resizebox{0.4\columnwidth}{!}{
\captionsetup{type=table}
\begin{tabular}{r|c}
\thickhline
\rowcolor{mygray}
\textbf{Method} & \textbf{Avg. Acc} \\ 
\hline
elect + mean   & \textbf{82.2} \\
elect + sum     & 79.6 \\
mean            & 79.7 \\
sum             & 79.8 \\
\thickhline
\end{tabular}
}
\caption{\label{tab:ablation} Comparison of $\psi(\cdot)$ on average accuracy.}
\end{table}

%% file: tables/ablation_adamerging.tex
\begin{table}[h]
\centering
\resizebox{0.85\linewidth}{!}{  
\begin{tabular}{r|c c}
\thickhline
\rowcolor{mygray}
\textbf{Method} & \textbf{ViT-B/32} & \textbf{ViT-L/14} \\ 
\hline
Layer-wise AdaMerging    & 80.1  & 90.8  \\
+NeuroMerging      & \textbf{82.3 \brc{(+2.2)}}  & \textbf{91.3 \brc{(+0.5)}} \\
\hline
Layer-wise AdaMerging++   & 81.1  & 91.0  \\
+NeuroMerging   & \textbf{82.5 \brc{(+1.4)}}  & \textbf{91.0 \brc{(+<0.1)}} \\
\thickhline
\end{tabular}
}
\caption{\label{tab:layer-wise-enhancement} Layer-wise enhancement on NeuroMerging.}
\end{table}

%% file: sections/7_Conclusion.tex
\section{Conclusion} 
In this paper, we revisited model merging from the core principle neuron connectivity and activation that underpin the learning process of recent deep neural networks and LLMs. Specifically, we presented the first exploration into the roles of neuronal mechanisms in the model merging process, by decomposing of task-specific representations into two complementary neuronal subspaces for characterization. It was mathematically shown that neuronal subspaces regulate input sensitivity and task adaptability. Based on these insights, we proposed NeuroMerging, a novel framework designed to reduce task interference within neurons. Our empirical evaluations demonstrated that NeuroMerging achieved superior performance compared to existing approaches on multi-task benchmarks across both vision and natural language processing tasks. 

Future work could extend NeuroMerging to larger models or multimodal architectures with more tasks. While our study introduced a neuronal mechanistic perspective for in model merging, it focused only on neuron connectivity and activation, without considering higher-order neuronal dynamics (e.g., network level connectivity and dynamics). Further research is needed to investigate these factors and deepen our understanding of neuronal mechanisms in model merging, or even in multi-task learning.

%% file: sections/8_Limitations.tex
\section*{Acknowledgements}
This work was supported in part by 
the Ministry of Higher Education Malaysia through the Fundamental Research Grant Scheme \seqsplit{(FRGS/1/2023/ICT02/XMU/02/1)}, 
National Science Foundation of China \seqsplit{(12204130,\ 62476070)}, 
Shenzhen Science and Technology Program \seqsplit{(JCYJ20241202123503005,\ GXWD20231128103232001)}, 
Department of Science and Technology of Guangdong \seqsplit{(2024A1515011540)}, 
Shenzhen Start-Up Research Funds \seqsplit{(HA11409065)}, 
and Xiamen University Malaysia Research Fund (XMUMRF/2024-C13/IECE/0049).

\section*{Limitations}
While our work provided the first neuronal mechanistic perspective on multi-task interference when merging large models, (1) it remains a partial view of neuronal mechanisms as it does not yet explore higher order neuronal dynamics during merging and inference, which require further investigation. Moreover, this work shares similar limitations with current SOTA model merging methods, including (2) the effectiveness of task arithmetic in model merging relies on selecting fine-tuned checkpoints that are beneficial for specific domains, ensuring they originate from the same pretrained model, and addressing hyperparameter sensitivity; and (3) More effort is needed to develop a mathematical understanding of why and when task arithmetic in model merging works well, despite its simplicity and efficiency.

\section*{Ethical Considerations}
Our research is based on publicly available datasets and models, all of which conform to their respective licenses and ethical guidelines. While the proposed NeuroMerging approach itself introduces no immediate ethical risks, caution should be exercised when generalizing results beyond the evaluated domains. Specifically, applying this technique to privacy-sensitive or high-stakes scenarios may require additional validation, as performance and reliability in untested contexts remain uncertain. We thus recommend thorough evaluation and responsible deployment practices in such applications.

%% file: sections/9_Appendix.tex
\clearpage
\appendix

\input{sections/Appendix/A_Algorithm}
\input{sections/Appendix/B_StatisticalAnalysis}

\input{sections/Appendix/C_AdditionalResults}
\input{sections/Appendix/D_Implementation}
\input{sections/Appendix/E_Dataset}

\clearpage
\input{tables/T5-Base}
\input{tables/ViT-Base}
\input{tables/ViT-Large}
\input{tables/T5-Base_OOD}
\input{tables/T5-Large_OOD}
\input{tables/intro_exp}
\begin{figure*}[t]
  \includegraphics[width=\linewidth]{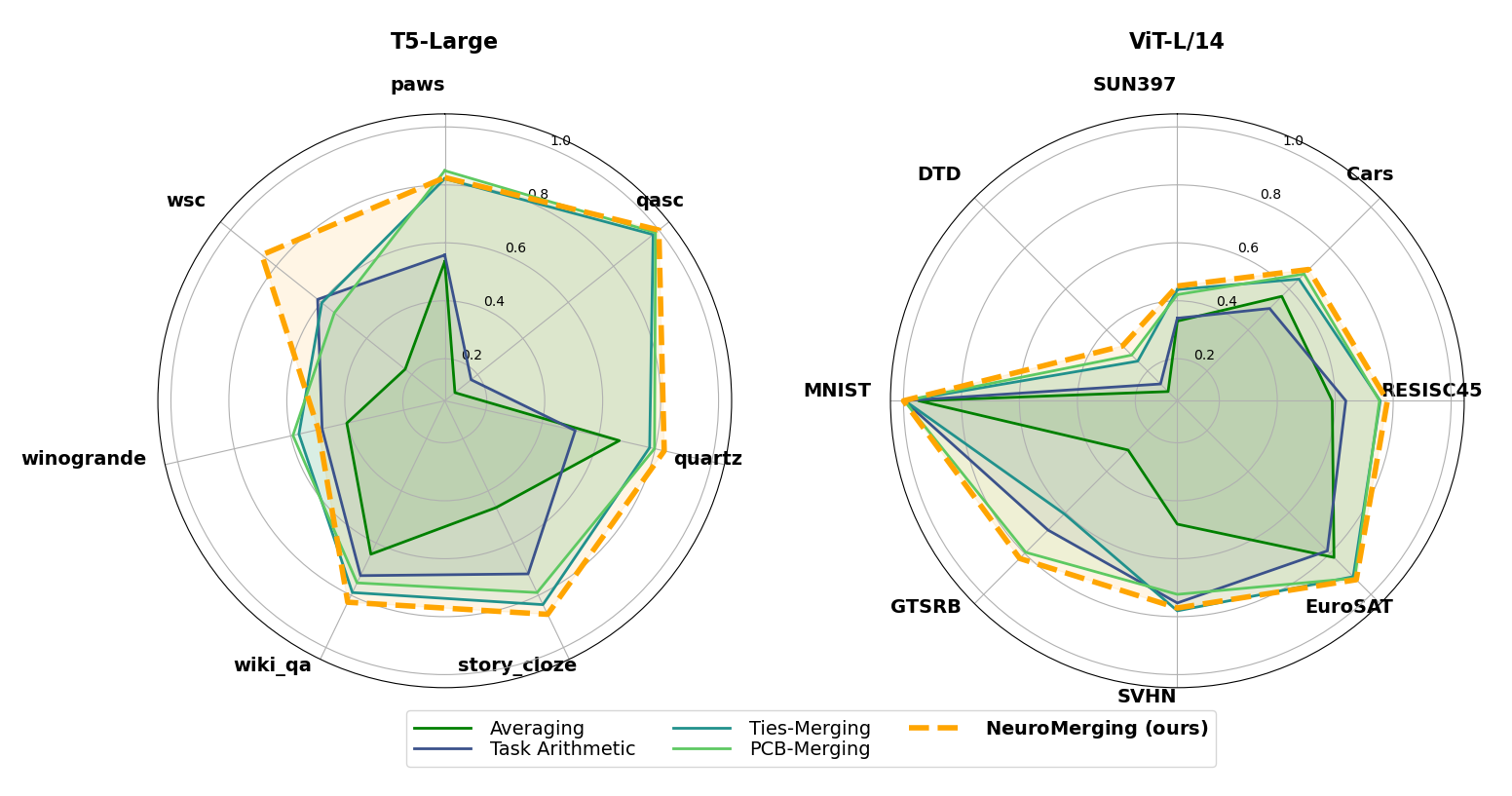}
  \caption {Comparison of merging methods on NLP with T5-Large \textbf{(Left)} and CV with ViT-L/14 \textbf{(Right)} without validation datasets. NeuroMerging outperformed existing methods in most tasks. Please refer to Section \ref{sec:Merging-without-Validation-Sets} for more discussion.}
  \label{fig:t5-large-vit-large-non-val}
\end{figure*}

%% file: sections/Appendix/A_Algorithm.tex
\section{Additional Details}
\label{sec:additional_details}

\subsection{Checkpoints for LLaMA}
We merged three domain-specialized Llama-2-7b models \citep{touvron2023llama}, and each is fine-tuned for distinct capabilities: 
Chinese language understanding\footnote{\url{https://huggingface.co/LinkSoul/Chinese-Llama-2-7b}}, 
mathematical reasoning\footnote{\url{https://huggingface.co/meta-math/MetaMath-7B-V1.0}} \citep{yu2024metamath}, 
and code generation\footnote{\url{https://huggingface.co/qualis2006/llama-2-7b-int4-python-code-18k}} \citep{roziere2023code}.

\subsection{Masking Ratio}
As entries of the task vector with larger magnitudes are more relevant to task-specific adaptation, only the top \( r\% \) of \( \mathbf{\tau}_t \) with the largest magnitudes are kept, while the others were set to zero~\cite{yadav2024ties}. The masked task vector is defined as \(  t \), where \( m_t \) is the mask that keeps the top \( r\% \) of the elements of each task vector. We used \( \tau_t \) to represent the masked task vector for readability.

\subsection{Singular Value Decomposition}
\label{sec:SVD}
We employed singular value decomposition (SVD) to construct a meaningful coordinate system within the orthogonal subspace $\mathcal{O}$, defined as the null space of the pre-trained neuron weights. Specifically, since $\mathcal{O}$ inherently lacks an explicit coordinate system, we first vertically stacked the task-specific neuron weight vectors $\tau_t^k$ into a matrix $D$, and then performed SVD on $D$ as follows:
\begin{equation}
    D = Q_k \Sigma_k P_k^\top
\end{equation}

Here, $k$ denotes the rank, which was set to be the number of tasks, and $P_k$ contains the top right singular vectors, which represent the dominant, i.e., principal, directions of task variability in the orthogonal subspace $\mathcal{O}$. These directions serve as the coordinate axes for the orthogonal subspace.

Next, we projected the task weights onto these axes using $DP_k$, and apply the merging function $\psi_\parallel(\cdot)$ column-wise to aggregate the projected components into a merged task representation $\zeta^k$ in the low-dimensional space. The final reconstruction of the merged weights was given by $\zeta^k P_k^\top$.

\subsection{Algorithm}
\label{sec:algo}
Algorithm~\ref{algo} is the pseudo-code for NeuroMerging.

\input{algorithms/NeuroMerging}

%% file: algorithms/NeuroMerging.tex
\begin{algorithm}[ht]
    \caption{NeuroMerging}
    \KwIn{Task-specific models $\tau_1, \tau_2, \dots, \tau_T$, pretrained model $\theta_0$, mask $m_t$, mask ratio $r$, proj matrix $\mathbf{P}$}
    \KwOut{Merged model $\bar{\theta}$}

    $\tau_t = m_t \circ \tau_t$ \tcp{Mask task vector based on $r$}

    \For{$k \gets 1$ \KwTo $K$}{
        \For{$t \gets 1$ \KwTo $T$}{
            \Comment{Create neuronal task vector.}
            $\tau^k_t = \mathbf{w}^k_t - \mathbf{w}^k_0$\;
            \BlankLine
            \Comment{Decompose neuronal subspaces.}
            $\tau^k_{\parallel, t} = \mathbf{P} \tau^k_t$, \hspace{0.1cm}
            $\tau^k_{\perp, t} = (\mathbf{I} - \mathbf{P}) \tau^k_t$\;
        }
    }
    
    \Comment{Merge neuronal task vectors.}

    \For{$k \gets 1$ \KwTo $K$}{
        \eIf{Validation data is available}{
            Tune $\lambda_1$ and $\lambda_2$ using the validation dataset\;
        }{
            $\sigma_t = \frac{\|\mathbf{\tau}_t^{masked}\|_1}{\|\mathbf{\tau}_t\|_1}$, \hspace{0.1cm}
            $\sigma = \max(\sigma_1, .., \sigma_T)$\;
            \BlankLine
            $\lambda_1 = 0$, \hspace{0.1cm} $\lambda_2 = \frac{1}{1 - \sigma}$\;
        }
        
        \small
        $\overline{\tau}^k = \lambda_1 \psi_{\parallel}(\tau^k_{\parallel, 1},.., \tau^k_{\parallel, T}) + \lambda_2 \psi_{\perp}(\tau^k_{\perp, 1}, .., \tau^k_{\perp, T})$\;
        \normalsize
    }
    \Comment{Reconstruct the merged task vector $\overline{\tau}$ by combining the $\overline{\tau}^k$ for each neuron.}
    
    $\bar{\theta} = \theta_0 + \overline{\tau}$ \hspace{0.2cm} \tcp{final merged model}
    \BlankLine
    
    \Return $\bar{\theta}$\;\label{algo}
\end{algorithm}

%% file: sections/Appendix/B_StatisticalAnalysis.tex
\section{Why Merging at the Neuronal Level?}
\label{sec:Statistical-Analysis-of-Parameters}

\begin{figure*}[ht]
  \centering
  \includegraphics[width=0.75\linewidth]{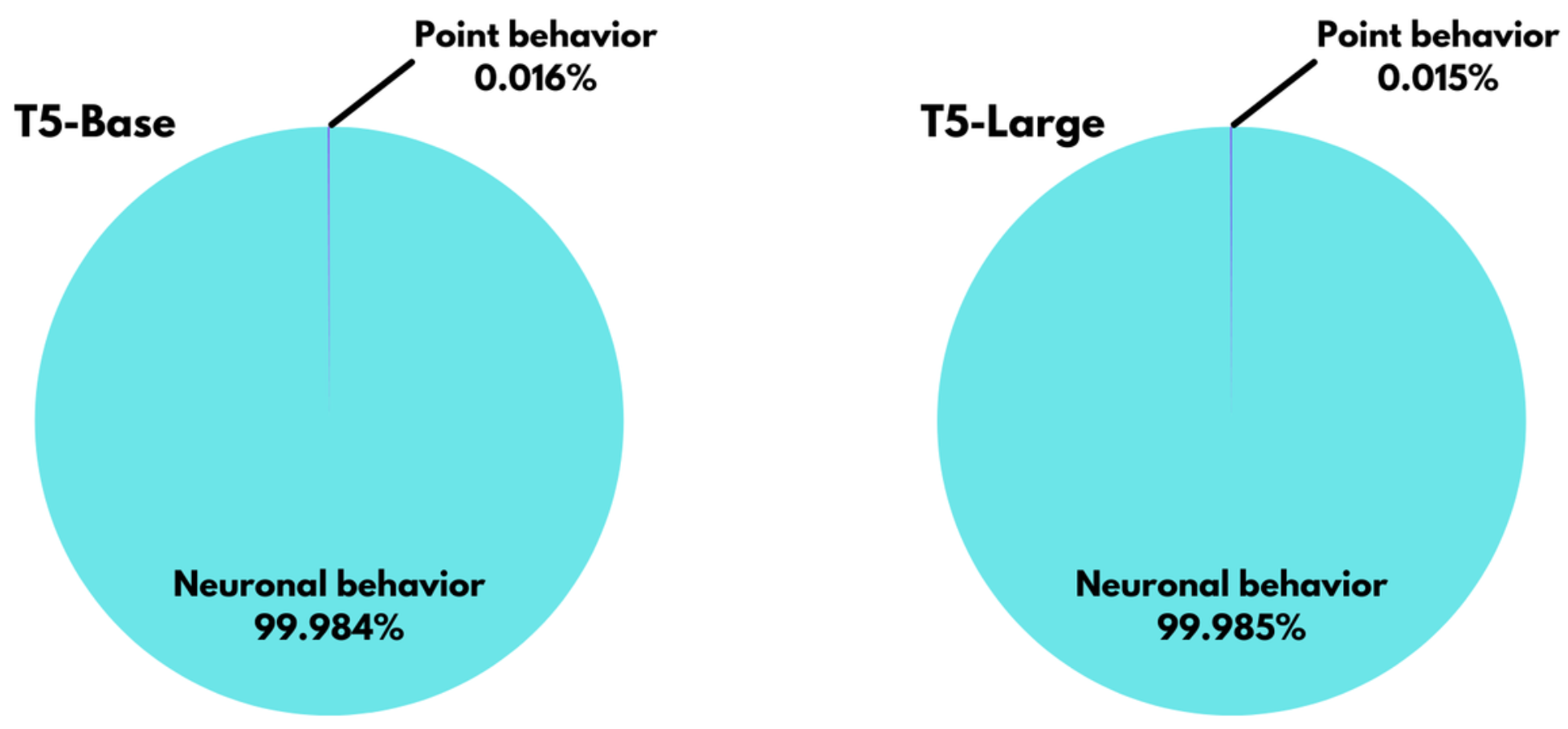}
  \caption{Statistical analysis of neuronal versus non-neuronal parameters.}
  \label{fig:parastats}
\end{figure*}

From the statistical analysis of neuronal versus non-neuronal parameters, it was observed that neuronal parameters dominate the T5-Base and T5-Large, showed in Figure.~\ref{fig:parastats}. As a result, they play a major role in shaping the model's learning dynamics, task adaptability, and overall performance. This dominance suggests that understanding and optimizing neuronal parameter interactions is crucial for improving model merging, reducing task interference, and enhancing generalization across diverse tasks. 

Moreover, we demonstrated in the main manuscript that merging in the orthogonal subspace ($\mathcal{O}$) is more effective. To explain why, we mathematically show that $\bar{x}_{\perp}$ does not influence the activation of the pre-trained neuron. For comparison, consider the activation of the pre-trained neuron:

\begin{equation}
    y = \phi(\mathbf{w}^k_0 \cdot \bar{x}) = \phi(\mathbf{w}^k_0 \cdot (\bar{x}_{\parallel} + \bar{x}_{\perp}))
\end{equation}

Since $\bar{x}_{\perp}$ belongs to the complementary orthogonal null space of $\mathbf{w}^k_0$, we have:

\begin{equation}
    \mathbf{w}^k_0 \cdot \bar{x}_{\perp} = 0
\end{equation}

Due to vector product of orthogonal subspaces, we simplify the activation as:

\begin{equation}
    y = \phi(\mathbf{w}^k_0 \cdot \bar{x}_{\parallel})
\end{equation}

Hence, the orthogonal subspace ($\mathcal{O}$) encapsulates more task-specific adaptations, making it a more effective space for merging. In Table~\ref{tab:diff-granularity}, we summarized the differences between our work with existing methods.


%% file: sections/Appendix/C_AdditionalResults.tex
\section{Additional Results}
\label{sec:Additional-Results}

\subsection{Comprehensive Task-Level Results}
\label{sec:Comprehensive-Task-Level-Results}
We provided all task-level results in T5-Base, T5-Large \citep{raffel2020exploring}, LLaMA2 \citep{touvron2023llama}, ViT-B/32, and ViT-L/14 \citep{dosovitskiy2021an}, respectively. The task-level results of the in-domain experiments for all models can be found in Tables~\ref{tab:t5-base},~\ref{tab:vit-b-32},~\ref{tab:vit-l-14}. The task-level results of the out-domain experiments for T5-Base and T5-Large can be found in Tables~\ref{tab:t5-base-ood} and~\ref{tab:t5-large-ood}. Lastly, Table \ref{tab:t5-large-intro-exp-all-table} showed the task-level results from Section~\ref{subspaceroles} when only one of the neuronal subspace was retained.

\subsection{Existing Model Merging Methods in Neuronal Subspaces}
To empirically validate the neuronal subspaces, we extended our experiments to various existing model merging methods. As shown in Tables~\ref{tab:nlp_under_Ortho_Para} and~\ref{tab:vision_under_Ortho_Para}, experimental results demonstrated that applying these merging methods within our proposed orthogonal neuronal subspaces consistently enhanced their performance. Notably, this improvement occurred despite these existing methods do not explicitly incorporating neuronal mechanisms into their designs. This observation further underscored the significance and motivation of our proposed approach.

\subsection{Scalability Evaluation}\label{sec:scalability_eval}
We further evaluated the scalability of NeuroMerging from the perspective of computational efficiency. As summarized in table~\ref{tab:nlp_efficiency} and~\ref{tab:vision_efficiency} (runtime in seconds, memory usage in GB; lower is better), NeuroMerging consistently achieved superior efficiency compared to existing methods across representative NLP and CV benchmarks. Specifically, it demonstrated comparable or reduced memory usage and improved accuracy relative to TIES-Merging~\cite{yadav2024ties} ($+2.7\%$ on T5-Large, $+2.3\%$ on ViT-L/14, $+1.7\%$ on LLaMA2), significantly lower runtime and memory overhead compared to PCB-Merging~\cite{du2024parameter}, and vastly reduced computational requirements relative to AdaMerging~\cite{yang2023adamerging}, which incured up to $100$ times longer runtime and $10$ times higher memory usage on ViT-L/14 due to additional training steps. In general, unlike prior approaches~\cite{yadav2024ties,du2024parameter,yang2023adamerging,jung2024tint,yang2024representation,huang2024emr,lu2024twin,matena2022merging}, which either required loading all parameters simultaneously or introduce costly auxiliary components, NeuroMerging leveraged neuronal task vectors aligned with neuronal mechanisms, ensuring scalable, lightweight, and efficient merging.

\subsection{Error Analysis}
\label{error_analysis}
This section analyzes individual sub-tasks where NeuroMerging's performance is more than $3\%$ lower compared to the best-performing baseline. The detailed breakdown covers NLP-series, Vision-series, and LLM-series tasks across various settings, including with and without validation, different model sizes, as well as in-domain and out-of-distribution (OOD) evaluations. For the quantitative analysis, we report both the performance differences and the single-task rankings. For the qualitative analysis, we offer possible explanations to interpret these observed differences.

\noindent\textbf{NLP series.} Performance details are summarized in Tables~\ref{tab:t5-large}, \ref{tab:t5-base}, \ref{tab:t5-base-ood}, and \ref{tab:t5-large-ood}. On Winogrande, T5-Large without validation is behind by $6.8\%$ (ranking 3/5), and T5-Base with validation by $6.6\%$ (ranking 6/6); on Quartz, T5-Base with and without validation lag by $6.3\%$ (ranking 4/6) and $3.3\%$ (ranking 3/5), respectively; on Story Cloze, T5-Base without validation trails by $5.5\%$ (ranking 2/5); for Cosmos QA in the out-of-distribution (OOD) setting, T5-Large is lower by $9.7\%$ (ranking 6/8); and on WiC, T5-Large and T5-Base under OOD conditions show performance gaps of $6.1\%$ (ranking 3/8) and $13.6\%$ (ranking 3/8), respectively.

These tasks typically require deeper semantic reasoning or more fine-grained contextual alignment compared to other tasks. Winogrande~\citep{sakaguchi2021winogrande} and WiC~\citep{pilehvar-camacho-collados-2019-wic} both rely heavily on nuanced lexical disambiguation or pronoun resolution. In these scenarios, the multi-task merging process, especially under limited model capacity or validation scenarios, may inadvertently ignore task-specific weights, diluting fine-grained semantic signals. Quartz~\citep{tafjord-etal-2019-quartz} and Cosmos QA~\citep{huang-etal-2019-cosmos} demand coherent integration of multi-sentence causal chains and background knowledge. However, parameter interference from shorter-text tasks occasionally compresses the reasoning chain, thus restricting deep semantic extraction. Story Cloze~\citep{sharma-etal-2018-tackling} emphasizes narrative coherence modeling, where different optimal configuration conflicts among merged tasks can weaken sensitivity to long-range dependencies. Overall, the performance gaps identified mainly arise from representational shifts induced by cross-task distributional differences, coupled with mismatches between model capacity and task complexity. These findings indicate promising directions for future research on model merging.

\noindent\textbf{Vision series.} Performance details are summarized in Tables~\ref{tab:vit-b-l}, \ref{tab:vit-b-32} and \ref{tab:vit-l-14}. On SUN397~\citep{2016SUN397}, ViT-B/32 with validation lag by $3.3\%$ (ranking 3/8); and on Cars~\citep{2013Cars}, ViT-B/32 with validation is behind by $3.5\%$ (ranking 2/8), with results similar to other merging methods except for Fisher Merging.

Fisher Merging demonstrated strong performance on these datasets, likely because it leverages the Fisher Information Matrix to selectively emphasize task-critical parameters on these two datasets (e.g., SUN397 requires panoramic scene composition, while Cars hinges on sub-class visual micro-signatures) through computing the Fisher Information Matrix using second-order derivatives or approximation through backpropagation. However, the high computational cost of the Fisher Information Matrix and its dependence on data make it less practical. As a result, recent model merging methods often avoid using the Fisher Information Matrix despite its effectiveness on these two datasets.

\noindent\textbf{LLM series.} Performance details are summarized in Table~\ref{tab:llama2}. NeuroMerging does not exhibit any sub-task with a performance drop exceeding $3\%$ in this scenario.

\input{tables/nlp_under_Ortho_Para}
\input{tables/vision_under_Ortho_Para}
\input{tables/computation_nlp}
\input{tables/computation_vision}

%% file: tables/nlp_under_Ortho_Para.tex
\begin{table*}[t!]
\centering
\captionsetup{type=table}
\small
\renewcommand{\arraystretch}{1.2}
\resizebox{0.85\textwidth}{!}{
\begin{tabular}{r|ccc|ccc}
\thickhline
\rowcolor{mygray}\textbf{Task($\rightarrow$)} &
\multicolumn{3}{c|}{\textbf{T5-Base}} & 
\multicolumn{3}{c}{\textbf{T5-Large}} \\
\cline{2-7}
\rowcolor{mygray}\textbf{Method($\downarrow$)} 
 & \textbf{Acc. (\%)} & \textbf{Orth. ($\mathcal{O}$)} & \textbf{Par. ($\mathcal{P}$)} 
 & \textbf{Acc. (\%)} & \textbf{Orth. ($\mathcal{O}$)} & \textbf{Par. ($\mathcal{P}$)} \\
\hline
TIES-Merging\pub{NeurIPS2023}      & \textbf{76.4} & 76.2 & 55.5 & 79.5 & \textbf{79.9} & 52.5 \\
PCB-Merging\pub{NeurIPS2024}       & 76.9 & \textbf{77.1} & 54.7 & 81.0 & \textbf{81.8} & 52.0 \\
\textbf{NeuroMerging (Ours)}       & \textbf{77.5} & \textbf{77.5} & 54.6 & \textbf{82.2} & \textbf{82.2} & 52.2 \\
\bottomrule
\end{tabular}
}
\caption{\label{tab:nlp_under_Ortho_Para}NLP tasks' accuracy under orthogonal and parallel projections across T5-Large and T5-Base models.}
\end{table*}

%% file: tables/vision_under_Ortho_Para.tex
\begin{table*}[t!]
\centering
\captionsetup{type=table}
\renewcommand{\arraystretch}{1.2}
\resizebox{0.85\textwidth}{!}{
\begin{tabular}{r|ccc|ccc}
\thickhline
\rowcolor{mygray}\textbf{Task($\rightarrow$)} &
\multicolumn{3}{c|}{\textbf{ViT-B/32}} & 
\multicolumn{3}{c}{\textbf{ViT-L/14}} \\
\cline{2-7}
\rowcolor{mygray}\textbf{Method($\downarrow$)} 
 & \textbf{Acc. (\%)} & \textbf{Orth. ($\mathcal{O}$)} & \textbf{Par. ($\mathcal{P}$)} 
 & \textbf{Acc. (\%)} & \textbf{Orth. ($\mathcal{O}$)} & \textbf{Par. ($\mathcal{P}$)} \\
\hline
Task-wise AdaMerging\pub{ICLR2024}        & 71.1 & \textbf{71.8} & 48.9 & 84.9 & \textbf{85.0} & 65.1 \\
Task-wise AdaMerging++\pub{ICLR2024}      & 73.7 & \textbf{74.5} & 49.2 & 87.3 & \textbf{87.5} & 65.6 \\
Layer-wise AdaMerging\pub{ICLR2024}        & 80.1 & \textbf{80.4} & 50.4 & 90.8 & \textbf{91.0} & 66.1 \\
Layer-wise AdaMerging++\pub{ICLR2024}      & \textbf{81.1} & \textbf{81.1} & 50.3 & 91.0 & \textbf{91.2} & 65.9 \\
PCB-Merging\pub{NeurIPS2024}         & 76.3 & \textbf{76.4} & 48.7 & 87.5 & \textbf{87.8} & 65.0 \\
\textbf{NeuroMerging (Ours)}         & 76.5 & \textbf{76.6} & 48.7 & \textbf{88.3} & \textbf{88.3} & 65.1 \\
\bottomrule
\end{tabular}
}
\caption{\label{tab:vision_under_Ortho_Para}Vision tasks' accuracy under orthogonal and parallel projections across ViT-B/32 and ViT-L/14 models.}
\end{table*}

%% file: tables/computation_nlp.tex
\begin{table*}[t!]
\centering
\captionsetup{type=table}
\renewcommand{\arraystretch}{1.2}
\resizebox{0.85\textwidth}{!}{
\begin{tabular}{r|ccc|ccc}
\thickhline
\rowcolor{mygray}\textbf{Task($\rightarrow$)} &
\multicolumn{3}{c|}{\textbf{T5-Base}} & 
\multicolumn{3}{c}{\textbf{T5-Large}} \\
\cline{2-7}
\rowcolor{mygray}\textbf{Method($\downarrow$)} 
 & \textbf{Time (s,$\downarrow$)} & \textbf{Mem (GB,$\downarrow$)} & \textbf{Acc. (\%,$\uparrow$)} 
 & \textbf{Time (s,$\downarrow$)} & \textbf{Mem (GB,$\downarrow$)} & \textbf{Acc. (\%,$\uparrow$)} \\
\hline
TIES-Merging\pub{NeurIPS2023}        & \textbf{17} & 61.4 & 76.4 & \textbf{50} & 190.1 & 79.5 \\
PCB-Merging\pub{NeurIPS2024}         & 98 & 105.5 & 76.9 & 323 & 312.1 & 81.0 \\
\textbf{NeuroMerging (Ours)}         & 67 & \textbf{18.7} & \textbf{77.5} & 161 & \textbf{69.1} & \textbf{82.2} \\
\bottomrule
\end{tabular}
}
\caption{\label{tab:nlp_efficiency}Comparison on computational efficiency and accuracy for NLP tasks.}
\end{table*}

%% file: tables/computation_vision.tex
\begin{table*}[t!]
\centering
\captionsetup{type=table}
\renewcommand{\arraystretch}{1.2}
\resizebox{0.85\textwidth}{!}{
\begin{tabular}{r|ccc|ccc}
\thickhline
\rowcolor{mygray}\textbf{Task($\rightarrow$)} &
\multicolumn{3}{c|}{\textbf{ViT-B/32}} & 
\multicolumn{3}{c}{\textbf{ViT-L/14}} \\
\cline{2-7}
\rowcolor{mygray}\textbf{Method($\downarrow$)} 
 & \textbf{Time (s,$\downarrow$)} & \textbf{Mem (GB,$\downarrow$)} & \textbf{Acc. (\%,$\uparrow$)} 
 & \textbf{Time (s,$\downarrow$)} & \textbf{Mem (GB,$\downarrow$)} & \textbf{Acc. (\%,$\uparrow$)} \\
\hline
TIES-Merging\pub{NeurIPS2023}        & \textbf{12} & 28.4 & 73.6 & \textbf{36} & 84.7 & 86.0 \\
PCB-Merging\pub{NeurIPS2024}         & 51 & 45.4 & 76.3 & 149 & 137.1 & 87.5 \\
AdaMerging\pub{ICLR2024}         & 6793 & 80.1 & 71.1 & 18871 & 265.4 & 84.9 \\
AdaMerging++\pub{ICLR2024}         & 6384 & 95.7 & 73.1 & 18184 & 335.7 & 87.3 \\
\textbf{NeuroMerging (Ours)}         & 67 & \textbf{8.8} & \textbf{76.5} & 141 & \textbf{21.7} & \textbf{88.3} \\
\bottomrule
\end{tabular}
}
\caption{\label{tab:vision_efficiency}Comparison on computational efficiency and accuracy for vision tasks.}
\end{table*}

%% file: sections/Appendix/D_Implementation.tex
\section{Implementation Details}
\label{sec:Implementation-Details}
We executed NLP and CV experiments on Nvidia GeForce 4090 GPUs with 24GB RAM, and LLM experiments on Nvidia A6000 GPUs with 48GB RAM, respectively.
The merging experiments demonstrated highly computational efficiency, with evaluation times under 2 minutes for T5-Base, T5-Large, ViT-B/32, and ViT-L/14 models. For large language model, specifically LLaMA2, the validation process across three datasets required approximately 40 minutes per complete evaluation~cycle. 

%% file: sections/Appendix/E_Dataset.tex
\section{Dataset Details}
\label{sec:Dataset-Details}
We utilized several datasets, and each of them comes with specific licenses. The following datasets are available under the Creative Commons License: 
\mbox{WiC}~\citep{pilehvar-camacho-collados-2019-wic}, 
\mbox{WSC}~\citep{levesque2012winograd}, 
Story Cloze~\citep{sharma-etal-2018-tackling}, 
\mbox{QuaRTz}~\citep{tafjord-etal-2019-quartz}, 
\mbox{Cars}~\citep{2013Cars}, 
and \mbox{GTSRB}~\citep{2011GTSRB}. 
\mbox{Winogrande}~\citep{sakaguchi2021winogrande} 
and \mbox{QASC}~\citep{Khot2019QASC} are distributed under the Apache License, 
while \mbox{COPA}~\citep{gordon-etal-2012-semeval} is covered by the BSD-2 Clause License. 
\mbox{WikiQA}~\citep{yang-etal-2015-wikiqa} is governed by the Microsoft Research Data License Agreement.
\mbox{Cosmos QA}~\citep{huang-etal-2019-cosmos} is licensed under the CC BY 4.0.
\mbox{QuAIL}~\citep{Rogers_Kovaleva_Downey_Rumshisky_2020} and \mbox{CMMLU}~\citep{li-etal-2024-cmmlu} are licensed under the CC BY-NC-SA 4.0.
\mbox{H-SWAG}~\citep{zellers-etal-2019-hellaswag},
\mbox{GSM8K}~\citep{cobbe2021training},
\mbox{HumanEval}~\citep{chen2021evaluating},
and \mbox{EuroSAT}~\citep{2019EuroSAT} fall under the MIT License, 
and \mbox{MNIST}~\citep{lecun1998mnist} is licensed under the GNU General Public License. 

For the datasets 
\mbox{DTD}~\citep{2014DTD}, 
\mbox{RESISC45}~\citep{2017RESISC45}, 
\mbox{SUN397}~\citep{2016SUN397}, 
\mbox{SVHN}~\citep{2011SVHN}, 
\mbox{Social IQA}~\citep{sap2019social},
and \mbox{PAWS}~\citep{paws2019naacl}, we were unable to determine specific licenses. 
However, they are publicly shared for research and education purposes.

%% file: tables/T5-Base.tex
\begin{table*}[t]
\centering
\captionsetup{type=table}
\resizebox{1.0\linewidth}{!}{  
\begin{tabular}{r|c|c|cccccccc}
\thickhline
\rowcolor{mygray}\textbf{Task($\rightarrow$)} &  &   & \multicolumn{7}{c}{\textbf{Test Set Performance}} \\ 
\cline{4-10}
\rowcolor{mygray}\textbf{Method($\downarrow$)} & \multirow{-2}{*}{\textbf{Validation}} & \multirow{-2}{*}{\textbf{Average}}  & paws & qasc & quartz & story\_cloze & wiki\_qa & winogrande & wsc \\ 
\hline
Zeroshot  & -  & 53.5  & 49.9  & 35.8  & 53.3  & 48.1  & 76.2  & 50.0  & 61.1 \\
Finetuned & -  & 79.6  & 93.9  & 98.0  & 81.4  & 82.5  & 95.4  & 51.9  & 54.2 \\
Multitask & -  & 83.6  & 94.0  & 97.9  & 82.5  & 86.7  & 95.0  & 64.1  & 65.3 \\
\hline
Averaging\pub{ICML22} & \ding{55}  & 64.2  & 65.1  & 81.1  & 59.9  & 48.8  & 94.7  & 50.9  & 48.6 \\
Task Arithmetic\pub{ICLR23}        & \ding{55}  & 60.6  & 78.5  & 30.0  & 56.8  & 65.6  & 95.0  & 49.6  & 48.6 \\
TIES-Merging\pub{NeurIPS23}      & \ding{55}  & \underline{73.6}  & \underline{82.4}  & \underline{94.1}  & \underline{71.9}  & 66.0  & \underline{91.2}  & \textbf{51.5}  & \underline{58.3} \\
PCB-Merging\pub{NeurIPS24}       & \ding{55}  & 73.4  & \textbf{83.1}  & 92.6  & \textbf{72.6}  & \textbf{73.4}  & 88.0  & \textbf{51.5}  & 52.8 \\
\textbf{NeuroMerging (Ours)} & \ding{55}  & \textbf{74.8}  & 81.8  & \textbf{96.5}  & 69.3  & \underline{67.9}  & \textbf{94.8}  & \underline{51.0}  & \textbf{62.5} \\
\hline
Fisher Merging\pub{NeurIPS22}   & \checkmark  & 68.3  & 66.7  & 85.6  & 63.5  & 57.1  & 90.1  & \underline{54.2}  & \underline{60.8} \\
RegMean\pub{ICLR23}  & \checkmark  & 72.7  & 77.2  & 93.8  & 63.6  & 64.6  & 90.4  & \textbf{58.4}  & 60.7 \\
Task Arithmetic\pub{ICLR23}       & \checkmark  & 73.6  & 83.2  & 89.9  & 69.3  & 72.9  & \textbf{95.2}  & 52.1  & 52.8 \\
TIES-Merging\pub{NeurIPS23}     & \checkmark  & 76.4  & \textbf{88.6}  & 94.1  & \textbf{74.5}  & 75.6  & 92.1  & 53.2  & 56.9 \\
PCB-Merging\pub{NeurIPS24}      & \checkmark  & \underline{76.9}  & \underline{88.2}  & \underline{95.2}  & \underline{71.0}  & \textbf{77.3}  & \underline{95.1}  & 51.9  & 59.7 \\
\textbf{NeuroMerging (Ours)}  & \checkmark  & \textbf{77.5}  & 87.5  & \textbf{95.7}  & 68.2  & \underline{76.8}  & 94.5  & 51.8  & \textbf{68.1} \\
\thickhline
\end{tabular}
}
\caption{\label{tab:t5-base} Test set performance when merging T5-Base models on seven NLP tasks. Please refer to Section \ref{sec:Merging-NLP-Models} for experimental details and Section \ref{error_analysis} for error analysis.}
\end{table*}

%% file: tables/ViT-Base.tex
\begin{table*}[t]
\centering
\captionsetup{type=table}
\resizebox{1.0\linewidth}{!}{  
\begin{tabular}{r|c|c|cccccccc}
\thickhline
\rowcolor{mygray}\textbf{Task($\rightarrow$)} &  &   & \multicolumn{8}{c}{\textbf{Test Set Performance}} \\ 
\cline{4-11}
\rowcolor{mygray}\textbf{Method($\downarrow$)} & \multirow{-2}{*}{\textbf{Validation}} & \multirow{-2}{*}{\textbf{Average}}  & SUN397 & Cars & RESISC45 & EuroSAT & SVHN & GTSRB & MNIST & DTD \\ 
\hline

Individual & -  & 90.5  & 75.3  & 77.7  & 96.1  & 99.7  & 97.5  & 98.7  & 99.7  & 79.4 \\
Multitask  & -  & 88.9  & 74.4  & 77.9  & 98.2  & 98.9  & 99.5  & 93.9  & 72.9  & 95.8 \\
\hline
Averaging\pub{ICML22}  & \ding{55}  & 65.8  & \underline{65.3}  & 63.4  & 71.4  & 71.7  & 64.2  & 52.8  & 87.5  & 50.1 \\
Task Arithmetic\pub{ICLR23}         & \ding{55}  & 60.4  & 36.7  & 41.0  & 53.8  & 64.4  & 80.6  & 66.0  & 98.1  & 42.5 \\
TIES-Merging\pub{NeurIPS23}       & \ding{55}  & 72.4  & 59.8  & 58.6  & 70.7  & 79.7  & \textbf{86.2}  & 72.1  & \underline{98.3}  & 54.2 \\
PCB-Merging\pub{NeurIPS24}        & \ding{55}  & \underline{75.9}  & \textbf{65.8}  & \textbf{64.4}  & \textbf{78.1}  & \underline{81.1}  & \underline{84.9}  & \underline{77.1}  & 98.0  & \underline{58.4} \\
\textbf{NeuroMerging (Ours)} & \ding{55}  & \textbf{76.4}  & 64.7  & \underline{64.2}  & \underline{77.0}  & \textbf{83.9}  & \textbf{86.2}  & \textbf{78.0}  & \textbf{98.5}  & \textbf{58.7} \\
\hline
Fisher Merging\pub{NeurIPS22}   & \checkmark  & 68.3  & \textbf{68.6}  & \textbf{69.2}  & 70.7  & 66.4  & 72.9  & 51.1  & 87.9  & \textbf{59.9} \\
RegMean\pub{ICLR23}  & \checkmark  & 71.8  & 65.3  & 63.5  & 75.6  & 78.6  & 78.1  & 67.4  & 93.7  & 52.0 \\
Task Arithmetic\pub{ICLR23}       & \checkmark  & 70.1  & 63.8  & 62.1  & 72.0  & 77.6  & 74.4  & 65.1  & 94.0  & 52.2 \\
TIES-Merging\pub{NeurIPS23}     & \checkmark  & 73.6  & 64.8  & 62.9  & 74.3  & 78.9  & 83.1  & 71.4  & 97.6  & 56.2 \\
AdaMerging\pub{ICLR2024}    &
\checkmark & 71.1 & 58.0 & 53.2
& 68.8 & \textbf{85.7} & 81.1 & \textbf{84.4} &
92.4 & 44.8 \\
AdaMerging++\pub{ICLR2024}    &
\checkmark & 73.7 & 60.8 & 56.9
& 73.1 & 83.4 & \textbf{87.3} & \underline{82.4} &
95.7 & 50.1 \\
PCB-Merging\pub{NeurIPS24}      & \checkmark  & \underline{76.3}  & \underline{66.7}  & 65.5  & \textbf{78.5}  & 79.3  & \underline{86.4}  & 77.1  & \underline{98.2}  & \underline{59.1} \\
\textbf{NeuroMerging (Ours)} & \checkmark  & \textbf{76.5}  & 65.3  & \underline{65.7}  & \underline{77.1}  & \underline{84.8}  & 84.5  & 77.9  & \textbf{98.3}  & 58.5 \\
\thickhline
\end{tabular}
}
\caption{\label{tab:vit-b-32} Test set performance when merging ViT-B/32 models on eight vision tasks. Please refer to Section \ref{sec:Merging-Vision-Models} for experimental details and Section \ref{error_analysis} for error analysis.}
\end{table*}

%% file: tables/ViT-Large.tex
\begin{table*}[t]
\centering
\captionsetup{type=table}
\resizebox{1.0\linewidth}{!}{  
\begin{tabular}{r|c|c|cccccccc}
\thickhline
\rowcolor{mygray}\textbf{Task($\rightarrow$)} &  &   & \multicolumn{8}{c}{\textbf{Test Set Performance}} \\ 
\cline{4-11}
\rowcolor{mygray}\textbf{Method($\downarrow$)} & \multirow{-2}{*}{\textbf{Validation}} & \multirow{-2}{*}{\textbf{Average}}  & SUN397 & Cars & RESISC45 & EuroSAT & SVHN & GTSRB & MNIST & DTD \\ 
\hline

Individual & -  & 94.2  & 82.3  & 92.4  & 97.4  & 100  & 98.1  & 99.2  & 99.7  & 84.1 \\
Multitask  & -  & 93.5  & 90.6  & 84.4  & 99.2  & 99.1  & 99.6  & 96.3  & 80.8  & 97.6 \\
\hline
Averaging\pub{ICML22}  & \ding{55}  & 79.6  & 72.1  & 81.6  & 82.6  & 91.9  & 78.2  & 70.7  & 97.1  & 62.8 \\
Task Arithmetic\pub{ICLR23}         & \ding{55}  & 83.3  & 72.5  & 79.2  & 84.5  & 90.6  & 89.2  & 86.5  & \underline{99.1}  & 64.3 \\
TIES-Merging\pub{NeurIPS23}       & \ding{55}  & 86.0  & \underline{76.5}  & 85.0  & \underline{89.3}  & 95.7  & \textbf{90.3}  & 83.3  & 99.0  & 68.8 \\
PCB-Merging\pub{NeurIPS24}        & \ding{55}  & \underline{86.9}  & 75.8  & \underline{86.0}  & 89.2  & \underline{96.0}  & 88.0  & \underline{90.9}  & \underline{99.1}  & \underline{70.0} \\
\textbf{NeuroMerging (Ours)} & \ding{55}  & \textbf{87.9}  & \textbf{77.0}  & \textbf{86.9}  & \textbf{90.3}  & \textbf{96.3}  & \underline{89.9}  & \textbf{92.1}  & \textbf{99.2}  & \textbf{71.8} \\
\hline
Fisher Merging\pub{NeurIPS22}   & \checkmark  & 82.2  & 69.2  & \textbf{88.6}  & 87.5  & 93.5  & 80.6  & 74.8  & 93.3  & 70.0 \\
RegMean\pub{ICLR23}  & \checkmark  & 83.7  & 73.3  & 81.8  & 86.1  & 97.0  & 88.0  & 84.2  & 98.5  & 60.8 \\
Task Arithmetic\pub{ICLR23}       & \checkmark  & 84.5  & 74.1  & 82.1  & 86.7  & 93.8  & 87.9  & 86.8  & 98.9  & 65.6 \\
TIES-Merging\pub{NeurIPS23}     & \checkmark  & 86.0  & 76.5  & 85.0  & \underline{89.4}  & 95.9  & \underline{90.3}  & 83.3  & \underline{99.0}  & 68.8 \\
AdaMerging\pub{ICLR2024}    &
\checkmark & 84.9 & 75.6 & 83.4
& 82.6 & 89.9 & 85.1 & \textbf{96.0} &
97.7 & 69.0 \\
AdaMerging++\pub{ICLR2024}    &
\checkmark & 87.3 & \underline{77.2} & 86.3
& 88.7 & 94.9 & 88.5 & \underline{93.2} & 98.8 & 71.3 \\
PCB-Merging\pub{NeurIPS24}      & \checkmark  & \underline{87.5}  & 76.8  & 86.2  & \underline{89.4}  & \textbf{96.5}  & 88.3  & 91.0  & 98.6  & \textbf{73.6} \\
\textbf{NeuroMerging (Ours)} & \checkmark  & \textbf{88.3}  & \textbf{77.3}  & \underline{87.1}  & \textbf{90.1}  & \underline{96.1}  & \textbf{91.0}  & 92.2  & \textbf{99.4}  & \underline{73.0} \\
\thickhline
\end{tabular}
}
\caption{\label{tab:vit-l-14} Test set performance when merging ViT-L/14 models on eight vision tasks. Please refer to Section \ref{sec:Merging-Vision-Models} for experimental details and Section \ref{error_analysis} for error analysis.}
\end{table*}

%% file: tables/T5-Base_OOD.tex
\begin{table*}[t]
\centering
\captionsetup{type=table}
\resizebox{1.0\linewidth}{!}{  
\begin{tabular}{r|c|cccccccc}
\thickhline
\rowcolor{mygray}
\textbf{Method ($\downarrow$)} & \textbf{Average} & cosmos\_qa & social\_iqa & quail & wic & copa & h-swag \\ 
\hline
paws         & 37.2  & 25.0  & 37.0  & 29.9  & 49.5  & 57.4  & 24.5 \\
qasc         & 36.5  & 21.2  & 37.4  & 29.5  & 49.8  & 54.4  & 26.7 \\
quartz       & 36.9  & 24.7  & 36.9  & 28.8  & 48.5  & 57.4  & 25.0 \\
story\_cloze & 36.8  & 21.9  & 36.4  & 25.7  & 53.6  & 57.4  & 26.1 \\
wiki\_qa     & 36.2  & 25.9  & 36.3  & 29.9  & 51.2  & 48.5  & 25.2 \\
winogrande   & 36.8  & 23.9  & 37.9  & 24.1  & 51.8  & 58.8  & 24.4 \\
wsc          & 39.5  & 26.9  & 38.1  & 29.5  & 55.4  & 61.8  & 25.2 \\
\hline
Pretrained   & 36.8  & 22.9  & 36.4  & 29.9  & 50.8  & 55.9  & 24.8 \\
Averaging\pub{ICML22}    & 37.4  & \textbf{23.7}  & 36.8  & 29.3  & 51.3  & 58.8  & 24.8 \\
Fisher Merging\pub{NeurIPS22}       & 33.8  & 15.6  & 21.9  & 24.9  & \textbf{65.6}  & 53.1  & 21.9 \\
Task Arithmetic\pub{ICLR23}           & 36.9  & 19.0  & 35.6  & \underline{29.5}  & \underline{54.0}  & 55.9  & \textbf{27.7} \\
RegMean\pub{ICLR23}      & 34.3  & \underline{23.1}  & 28.1  & 24.9  & 48.4  & 62.5  & 18.8 \\
TIES-Merging\pub{NeurIPS23}         & \underline{38.5}  & 21.9  & \underline{37.4}  & 29.3  & 52.0  & \underline{64.7}  & \underline{25.5} \\
PCB-Merging\pub{NeurIPS24}          & \underline{38.5}  & 22.8  & \textbf{37.5}  & 29.1  & 51.3  & 63.2  & 27.0 \\
\textbf{NeuroMerging (Ours)} & \textbf{39.2}  & 21.2  & 37.3  & \textbf{29.9}  & 52.0  & \textbf{69.1}  & 25.4 \\
\thickhline
\end{tabular}
}
\caption{\label{tab:t5-base-ood} Out-of-Distribution performance of T5-Base models checkpoints on six tasks. Please refer to Section~\ref{sec:Out-of-Domain-Generalization} for experimental details and Section \ref{error_analysis} for error analysis.}
\end{table*}

%% file: tables/T5-Large_OOD.tex
\begin{table*}[t]
\centering
\captionsetup{type=table}
\resizebox{1.0\linewidth}{!}{  
\begin{tabular}{r|c|cccccccc}
\thickhline
\rowcolor{mygray}
\textbf{Method ($\downarrow$)} & \textbf{Average} & cosmos\_qa & social\_iqa & quail & wic & copa & h-swag \\ 
\hline
paws         & 38.2  & 28.4  & 37.6  & 25.4  & 60.9  & 51.5  & 25.2 \\
qasc         & 37.9  & 23.1  & 37.0  & 25.5  & 49.0  & 64.7  & 28.1 \\
quartz       & 36.2  & 26.1  & 38.0  & 25.7  & 51.3  & 50.0  & 26.2 \\
story\_cloze & 37.9  & 22.9  & 37.5  & 24.5  & 51.2  & 64.7  & 26.6 \\
wiki\_qa     & 35.0  & 23.2  & 37.4  & 26.1  & 51.2  & 47.1  & 25.1 \\
winogrande   & 36.1  & 25.2  & 39.6  & 24.1  & 51.3  & 50.0  & 26.4 \\
wsc          & 37.2  & 26.2  & 38.8  & 28.8  & 55.4  & 48.5  & 25.8 \\
\hline
Pretrained   & 36.3  & 23.7  & 37.8  & 28.1  & 51.2  & 51.5  & 25.5 \\
Averaging\pub{ICML22}    & 36.7  & 25.3  & 37.0  & 23.4  & 51.5  & 57.4  & 25.9 \\
Fisher Merging\pub{NeurIPS22}       & 32.0  & \textbf{34.4}  & 25.0  & 26.1  & 40.6  & 56.2  & 9.4 \\
Task Arithmetic\pub{ICLR23}           & 39.2  & 24.6  & 38.0  & \textbf{27.3}  & \underline{58.6}  & 58.8  & 28.1 \\
RegMean\pub{ICLR23}      & 36.0  & \textbf{34.4}  & 28.1  & 25.3  & \textbf{62.5}  & 50.0  & 15.6 \\
TIES-Merging\pub{NeurIPS23}         & 40.1  & 25.1  & \textbf{40.8}  & 23.0  & 56.3  & \underline{67.6}  & 27.6 \\
PCB-Merging\pub{NeurIPS24}          & \underline{40.4}  & \underline{25.6}  & \underline{40.7}  & 25.7  & 55.1  & 66.2  & \textbf{29.3} \\
\textbf{NeuroMerging (Ours)} & \textbf{40.9}  & 24.7  & \underline{40.7}  & 26.6  & 56.4  & \textbf{69.1}  & \underline{27.9} \\
\thickhline
\end{tabular}
}
\caption{\label{tab:t5-large-ood} Out-of-Distribution performance of T5-Large models checkpoints on six tasks. Please refer to Section~\ref{sec:Out-of-Domain-Generalization} for experimental details and Section \ref{error_analysis} for error analysis.}
\end{table*}

%% file: tables/intro_exp.tex
\begin{table*}[t]
\centering
\captionsetup{type=table}
\resizebox{1.0\linewidth}{!}{  
\begin{tabular}{r|c|ccccccc|c}
\thickhline
\rowcolor{mygray}\textbf{Task($\rightarrow$)} &  & \multicolumn{7}{c}{\textbf{Test Set Performance}} &  \\ 
\cline{3-10}
\rowcolor{mygray}\textbf{Method($\downarrow$)} & \multirow{-2}{*}{\textbf{Dataset ($\downarrow$)}} & paws & qasc & quartz & story\_cloze & wiki\_qa & winogrande & wsc & \textbf{Average} \\ 
\hline

\multirow{7}{*}{\textbf{Fine-tuned}} 
& paws         & \cellcolor{mygray}94.4  & 18.0  & 53.3  & 53.1  & 87.9  & 49.8  & 54.2  & 58.7 \\
& qasc         & 54.3  & \cellcolor{mygray}97.1  & 55.7  & 64.7  & 66.3  & 49.8  & 41.7  & 61.4 \\
& quartz       & 59.9  & 65.1  & \cellcolor{mygray}85.3  & 51.2  & 72.5  & 48.9  & 62.5  & 63.6 \\
& story\_cloze & 53.4  & 31.9  & 54.2  & \cellcolor{mygray}91.0  & 57.4  & 49.1  & 56.9  & 56.3 \\
& wiki\_qa     & 55.8  & 16.3  & 50.9  & 53.7  & \cellcolor{mygray}95.7  & 48.7  & 63.9  & 55.0 \\
& winogrande   & 55.7  & 50.2  & 62.4  & 55.3  & 78.4  & \cellcolor{mygray}71.6  & 56.9  & 61.5 \\
& wsc          & 55.8  & 16.3  & 57.7  & 48.5  & 73.3  & 47.4  & \cellcolor{mygray}80.6  & 54.2 \\

\rowcolor{myyellow}
& \textbf{Total Avg.} & \multicolumn{2}{c}{\textbf{In-Domain: 88.0}} & \multicolumn{2}{c}{\textbf{Out-Domain: 53.8}} & \multicolumn{2}{c}{\textbf{All: 58.7}} & & \\

\hline
\multirow{7}{*}{\textbf{Orthogonal}} 
& paws         & \cellcolor{mygray}94.4  & 18.1  & 53.3  & 53.3  & 88.1  & 49.6  & 56.9  & 59.1 \\
& qasc         & 54.4  & \cellcolor{mygray}97.1  & 55.6  & 64.8  & 66.3  & 50.0  & 43.1  & 61.6 \\
& quartz       & 60.0  & 65.0  & \cellcolor{mygray}85.3  & 51.5  & 72.5  & 48.5  & 63.9  & 63.8 \\
& story\_cloze & 53.3  & 31.8  & 53.8  & \cellcolor{mygray}90.9  & 57.7  & 49.0  & 56.9  & 56.2 \\
& wiki\_qa     & 55.8  & 16.3  & 51.1  & 53.5  & \cellcolor{mygray}95.7  & 48.2  & 63.9  & 54.9 \\
& winogrande   & 55.7  & 49.8  & 62.2  & 55.8  & 78.4  & \cellcolor{mygray}71.7  & 56.9  & 61.5 \\
& wsc          & 55.8  & 16.6  & 57.4  & 48.2  & 73.4  & 47.4  & \cellcolor{mygray}80.6  & 54.2 \\

\rowcolor{myyellow}
& \textbf{Total Avg.} & \multicolumn{2}{c}{\textbf{In-Domain: 88.0}} & \multicolumn{2}{c}{\textbf{Out-Domain: 53.9}} & \multicolumn{2}{c}{\textbf{All: 58.8}} & & \\

\hline
\multirow{7}{*}{\textbf{Parallel}} 
& paws         & \cellcolor{mygray}54.4  & 14.5  & 53.4  & 54.2  & 71.8  & 49.5  & 63.9  & 51.7 \\
& qasc         & 55.3  & \cellcolor{mygray}14.9  & 54.0  & 54.3  & 71.0  & 48.6  & 63.9  & 51.7 \\
& quartz       & 55.3  & 14.3  & \cellcolor{mygray}55.6  & 54.2  & 71.7  & 49.4  & 63.9  & 52.1 \\
& story\_cloze & 55.3  & 14.3  & 54.1  & \cellcolor{mygray}53.8  & 71.1  & 49.3  & 63.9  & 51.7 \\
& wiki\_qa     & 55.6  & 14.7  & 54.5  & 54.3  & \cellcolor{mygray}77.1  & 49.1  & 63.9  & 52.7 \\
& winogrande   & 55.3  & 14.8  & 54.2  & 53.6  & 71.8  & \cellcolor{mygray}49.5  & 63.9  & 51.9 \\
& wsc          & 55.3  & 14.7  & 53.7  & 53.7  & 72.5  & 49.5  & \cellcolor{mygray}63.9  & 51.9 \\

\rowcolor{myyellow}
& \textbf{Total Avg.} & \multicolumn{2}{c}{\textbf{In-Domain: 52.7}} & \multicolumn{2}{c}{\textbf{Out-Domain: 51.8}} & \multicolumn{2}{c}{\textbf{All: 51.9}} & & \\

\thickhline
\end{tabular}
}
\caption{\label{tab:t5-large-intro-exp-all-table} Test set performance comparison of T5-Large models under different keeping strategies (naive finetuned, keep orthogonal, and keep parallel) across seven NLP tasks. Please refer to Section \ref{Introduction} and \ref{subspaceroles} for experimental details.}
\end{table*}